\documentclass[10pt,twocolumn,letterpaper]{article}

\usepackage{wacv}
\usepackage{times}
\usepackage{epsfig}
\usepackage{graphicx}
\usepackage{amsmath,bm}
\usepackage{amssymb}
\usepackage{booktabs}
\usepackage{subfloat}
\usepackage{caption}
\usepackage{subcaption}
\usepackage{float}
\usepackage{multirow}
\usepackage{multicol}
\usepackage{algorithm, algorithmic}

\usepackage[accsupp]{axessibility}

\def \x{\mathbf{x}}

\def\wacvPaperID{1080} % Enter the WACV Paper ID here

\wacvfinalcopy % *** Uncomment this line for the final submission

\ifwacvfinal
\usepackage[breaklinks=true,bookmarks=false]{hyperref}
\else
\usepackage[pagebackref=true,breaklinks=true,colorlinks,bookmarks=false]{hyperref}
\fi

\begin{document}

\pagenumbering{gobble}

%%%%%%%%% TITLE
\title{Adaptive Sample Selection for Robust Learning under Label Noise}

\author{Deep Patel and P S Sastry\\
	Indian Institute of Science, \\
	Bangalore, India - 560012\\
	{\tt\small deeppatel, sastry@iisc.ac.in}
}

\maketitle
\thispagestyle{empty}

\vspace*{-0.5cm}

%%%%%%%%% ABSTRACT
\begin{abstract}
	
Deep Neural Networks (DNNs) have been shown to be susceptible to memorization or overfitting in the presence of noisily-labelled data. For the problem of robust learning under such noisy data, several algorithms have been proposed. A prominent class of algorithms rely on sample selection strategies wherein, essentially, a fraction of samples with loss values below a certain threshold are selected for training. These algorithms are sensitive to such thresholds, and it is difficult to fix or learn these thresholds. Often, these algorithms also require information such as label noise rates which are typically unavailable in practice. In this paper, we propose an adaptive sample selection strategy that relies only on batch statistics of a given mini-batch to provide robustness against label noise. The algorithm does not have any additional hyperparameters for sample selection, does not need any information on noise rates and does not need access to separate data with clean labels. We empirically demonstrate the effectiveness of our algorithm on benchmark datasets.\footnote{Codes for reproducibility will be made available here: \url{https://github.com/dbp1994/masters_thesis_codes/tree/main/BARE}}
\end{abstract}

\vspace*{-0.6cm}
\section{Introduction}
\vspace*{-0.1cm}

%Label noise is inevitable when employing supervised learning methods. 
The deep learning models, which are highly effective in many applications, need vast amounts of training data. Such large-scale labelled data is often generated through crowd-sourcing or automated labeling, 
%(e.g., web crawlers to collect image data on the internet). 
which naturally cause random labelling errors. In addition, subjective biases in human annotators too can cause such errors. The training of deep networks is adversely affected by label noise and hence robust learning under label noise is an important problem of current interest.

In recent years many different approaches for robust learning of classifiers have been proposed, such as, robust loss functions \cite{ghosh2017aaai,sugiyama-sym-loss,gce,sym-ce}, loss correction \cite{loss-correction}, meta-learning \cite{meta-mlnt,meta-lc}, sample reweighting \cite{meta-ren,meta-net,joint-opt,mentornet,coteaching}, etc. In this paper we present a novel algorithm that adaptively selects samples based on the statistics of observed loss values in a minibatch and achieves good robustness to label noise. 
%Unlike other algorithms that rely on sample reweighting, 
Our algorithm does not use any additional system for learning weights for examples, does not need extra   data with clean labels and does not assume any knowledge of noise rates. 
%It is simple to implement and has no additional hyperparameters for the sample selection strategy. 
The algorithm is motivated by curriculum learning and can be thought of as a way to design an adaptive curriculum. 

\vspace*{-0.04cm}
The curriculum learning  \cite{bengio2009curriculum,spl} is a general strategy of sequencing of examples so that the networks learn from the `easy' examples well before learning from the `hard' ones. This is often brought about by giving different weights to different examples in the training set.
%often based on the values of loss function on these examples. 
%In the context of label noise, one can think of clean examples as the easy ones and the examples with wrong labels as the hard ones. 
Many of the recent algorithms for robust learning based on sample reweighting can be seen as motivated by a similar idea. 
%In all such approaches, the weight assigned to an example is essentially determined by the loss function value on that example with a heuristic that, low loss values indicate reliable labels. 
%Many different ways of fixing/learning such weights have been proposed (e.g., \cite{meta-ren,deep-bilevel,meta-net}) with the general heuristic being that low loss values indicate reliable labels. 
A good justification for this approach comes from some recent studies 
%on the effects of noisily-labelled data on learning 
%deep neural networks. It is empirically 
 \cite{zhang} that have shown that deep networks can learn to achieve zero training error on completely randomly labelled data, a phenomenon termed as `memorization'. Further studies such as \cite{arpit,d2l} have shown that the networks, when trained on noisily-labelled data, learn simpler patterns 
%(corresponding to cleanly-labelled data) 
first before overfitting to the noisily-labelled data.

In the last few years, several strategies have been proposed that aim to select (or give more weightage to) `clean' samples for obtaining a degree of robustness against label noise (e.g., \cite{mentornet, coteaching, coteaching+, s2e, curr-loss, meta-ren,deep-bilevel}). All such methods essentially employ the heuristic of `small loss' for sample selection wherein (a fraction of) small-loss valued samples are preferentially used for learning the network. Many of these methods use an auxiliary network to assess the loss of an example or to learn to map loss values to sample weights.
%  Algorithms such as Co-Teaching \cite{coteaching} and Co-Teaching+ \cite{coteaching+} use two networks and select samples with loss value below a threshold in one network to train the other. In Co-Teaching, the threshold is chosen based on the knowledge of noise rates and is fixed throughout the training. The same threshold is used in Co-Teaching+ but the sample selection is based on disagreement between the two networks. MentorNet \cite{mentornet}, another recent algorithm based on curriculum learning, uses an auxiliary neural network trained to serve as a sample selection function. There are other adaptive sample reweighting schemes based on meta-learning such as \cite{meta-ren, meta-net} which learn sample weights for the examples. % and they require some extra data with clean labels for learning these weights. 
%Algorithms in \cite{meta-net,mentornet} use a separate network to learn a mapping from loss values to weights of examples. 
Such methods need additional computing resources to learn multiple networks and may also need separate clean data (without label noise) and the methods involve careful choice of additional hyperparameters. 
%In addition, all such methods, in effect, assume that one can assess whether or not an example has clean label based on some function of the loss value of that example. 
%In addition, since these algorithms rely on a function of current loss value of an example to  decide whether or not the example is correctly labelled, they, in effect,  assume such a mapping exists. 
In general, it is difficult to directly relate the loss value of an example with the reliability of its label. 
Loss value of any specific example is itself a function of the current state of learning and it evolves with epochs. Loss values of even clean samples may change over a significant range during the course of learning.  Further, the loss values achievable by a network even on clean samples may be different for examples of different classes. 
%Thus, it is desirable to have a sample selection strategy that is based on assessing relative loss values of the set of examples rather than on thresholds on absolute loss values. 
% Methods based on curriculum learning such as \cite{coteaching,coteaching+} use a threshold to pick up the examples with small loss values in each mini-batch. This threshold is fixed using the noise rate which is not known and is often difficult to estimate reliably. (As we show in this paper, the method is somewhat sensitive to errors in estimated noise rate). While they adapt this threshold with epochs in a fixed manner, it is not really dependent on the current state of learning.
% as learning progresses, the adaptation is essentially a pre-designed schedule and needs some hyperparameters. 
%It is desirable that assessing of whether an example has clean label based on the loss value depends on the current state of learning. 

Motivated by these considerations, we propose a simple, adaptive 
%curriculum based 
selection strategy called \textit{BAtch REweighting} (\textbf{BARE}). 
%The idea is to focus on the current state of learning, in a given mini-batch, for identifying the noisily labelled data in it. 
%The statistics of loss values of all examples in a mini-batch would give useful information on current state of learning. 
Our algorithm utilizes the statistics of loss values in a mini-batch to compute the threshold for sample selection in that mini-batch. 
%This will give us what is essentially a dynamic or adaptive curriculum where the sample selection is naturally tied to state of learning. For example, 
Since, it is possible that this automatically calculated threshold is different for different mini-batches even within the same epoch, our method amounts to using a dynamic threshold which naturally evolves as learning proceeds. In addition, while calculating the batch statistics we take into consideration the class labels also and hence the dynamic thresholds are also dependent on the given labels of the examples. 

The main contribution of this paper is an adaptive sample selection strategy for robust learning that is simple to implement, does not need any clean validation data, needs no knowledge at all of the noise rates and also does not have any hyperparameters in the sample selection strategy. It does not need any auxiliary network for sample selection.   We empirically demonstrate the effectiveness of our algorithm on benchmark datasets: MNIST \cite{mnist}, CIFAR-10 \cite{cifar10}, and Clothing-1M \cite{clothing1M} and show that our algorithm is much more efficient in terms of time and has as good or better robustness compared to other algorithms  for different types of label noise and noise rates.

The rest of the paper is organized as follows: Section \ref{sec:lit-review} discusses related work, Section \ref{sec:bare} discusses our proposed algorithm. Section \ref{sec:exp} discusses our empirical results and concluding remarks are provided in Section \ref{sec:conclusion}.

\vspace*{-0.2cm}
\section{Related Work}
\label{sec:lit-review}

\vspace*{-0.15cm}
Curriculum learning (CL) as proposed in \cite{bengio2009curriculum} is the designing of an (optimal) manner of sequencing of training samples (based on a notion of \textit{easiness} of an example)  to improve the performance of the learning system. A curriculum called Self-Paced Learning (SPL) is proposed in \cite{spl} wherein easiness is decided upon based on how small the loss values are. A framework to unify CL and SPL is proposed in \cite{spcl}. SPL with diversity \cite{spld} proposes a sample selection scheme to encourage selection of a diverse set of \textit{easy} examples. This is further improved in \cite{minmax-cl} by encouraging more exploration during early phases of learning.  More recently, \cite{adaptive-curr-2021} propose a curriculum which computes exponential moving averages of loss values as difficulty scores for training samples.

Motivated by similar ideas, many sample reweighting algorithms are proposed for tackling label noise in neural networks. Many different ways of fixing/learning such weights have been proposed (e.g., \cite{mentornet, coteaching, coteaching+, s2e, curr-loss,meta-ren,deep-bilevel,meta-net}) with the general heuristic being that low loss values indicate reliable labels. Algorithms such as Co-Teaching \cite{coteaching} and Co-Teaching+ \cite{coteaching+} use two networks and select samples with loss value below a threshold in one network to train the other. In Co-Teaching, the threshold is chosen based on the knowledge of noise rates. The same threshold is used in Co-Teaching+ but the sample selection is based on disagreement between the two networks. \cite{curr-loss} also relies on `small loss' heuristic but the threshold for sample selection is adapted based on the knowledge of label noise rates.  MentorNet \cite{mentornet}, another recent algorithm based on curriculum learning, uses an auxiliary neural network trained to serve as a sample selection function.  Another sample selection algorithm is proposed in \cite{malach2017decoupling} where the idea is to train two networks and update the network parameters only in case of a disagreement between the two networks.  These sample selection functions are mostly hand-crafted and, hence, they can be sub-optimal. Another strategy is to solve a bilevel optimization problem to find the optimal sample weights (e.g., \cite{deep-bilevel}). The sample selection function used in \cite{coteaching, coteaching+} is sub-optimally chosen for which \cite{s2e} proposes an AutoML-based approach to find a better function, by fine-tuning on separate data with clean labels. Sample reweighting algorithms proposed in \cite{meta-ren} and \cite{meta-net} use online meta-learning and need some extra data with clean labels. 

\vspace*{-0.11cm}
Apart from the sample selection/reweighting approaches described above, there are other approaches to tackling label noise. \textit{Label cleaning} algorithms \cite{joint-opt, pencil, selfie} attempt at identifying and correcting the potentially incorrect labels through joint optimization of sample weights and network weights. \textit{Loss correction} methods \cite{loss-correction,meta-lc} suitably modify loss function (or posterior probabilities) to correct for the effects of label noise on risk minimization; however, they need to know (or estimate) the noise rates. There are also theoretical results that investigate robustness of risk minimization \cite{ghosh2017aaai, gce, rll, sym-ce, ma2020normalized, manwani2013}. \textit{Regularization} methods, of which sample reweighting approaches are a part, employ explicit or implicit regularization to reduce overfitting to noisy data \cite{unsup-loss-corr, dividemix, dim-driv-learn, zhang2017mixup, reed2015training, vat}.  More recently, some works have used \textit{self-supervised learning} methods to obtain better initializations for robustness \cite{aritra-self-2021, contrast-self-2022}, second-order statistics for label cleaning \cite{zhu2021second} and cluster-based consensus methods \cite{zhu2021clusterability} to improve noise transition matrix estimations thereby improving loss-correction methods. In this paper, our interest is in the approach of sample selection for achieving robustness to label noise.

%\vspace*{-0.05cm}
The proposed algorithm, BARE, is a simple, adaptive method to select samples which relies only on statistics of loss values (or, equivalently, statistics of class posterior probabilities because we use CCE loss) in a given mini-batch. We do not need any extra data with clean labels or any knowledge about label noise rates. Since it uses batch statistics, the selection thresholds are naturally tied to the evolving state of learning of the network without needing any tunable hyperparameters. Unlike in many of the aforementioned algorithms, we do not need any auxiliary networks for learning sample selection function, or cross-training, or noise rate estimation and, thus, our algorithm is computationally more efficient.

\vspace*{-0.25cm}

\section{Batch Reweighting Algorithm}
\label{sec:bare}
\vspace*{-0.1cm}
In this section we describe the proposed sample reweighting algorithm that relies on mini-batch statistics. 

\vspace*{-0.25cm}
\subsection{Problem Formualtion and Notation}
\label{sec:notation}

Under label noise, the labels provided in the training set may be `wrong' and we want a classifier whose test error with respect to `correct' labels is good. 
%We begin by making this more precise and introducing our notation.

Consider a $K$-class problem with $\mathcal{X}$ as the feature/pattern space and  $\mathcal{Y} = \{0,1\}^K $ as the label space. We assume all labels are one-hot vectors and denote by $e_k$ the one-hot vector corresponding to class $k$. Let $S^c = \{(x_i, y^c_i), \; i=1, 2, \cdots, m\}$ be iid samples drawn according to a distribution $\mathcal{D}$ on $\mathcal{X} \times \mathcal{Y}$. We are interested in learning a classifier that does well on a test set drawn according to $\mathcal{D}$. We can do so if we are given $S^c$ as training set. However, what we have is a training set $S=  \{(x_i, y_i), \; i=1, 2, \cdots, m\}$ drawn according to a distribution $\mathcal{D}_{\eta}$. The $y_i$ here are the `corrupted' labels and they are related to $y^c_i$, the `correct' labels through %a probabilistic relation
\begin{equation}
	P[y_i = e_{k'} \; | \; y_i^c = e_{k}] = \eta_{kk'}
\end{equation}
The $\eta_{kk'}$ are called noise rates. (In general the above probability can also depend on the feature vector, $x_i$, though we do not consider that possibility in this paper). We call this general model as class conditional noise because here the probability of label corruption depends on the original label. A special case of this is the so called symmetric noise where we assume $\eta_{kk}= (1 - \eta)$ and $\eta_{kk'} = \frac{\eta}{K-1}, \forall k' \neq k$. Here, $\eta$ represents the probability of a `wrong' label. With symmetric noise, the corrupted label is equally likely to be any other label. 

We can represent $\eta_{kk'}$ as a matrix and we assume it is diagonally dominant (that is, $\eta_{kk} > \eta_{kk'}, \forall k' \neq k$). (Note that this is true for symmetric noise if $\eta < \frac{K-1}{K}$). Under this condition, if we take all patterns labelled by a specific class in the label-corrupted training set, then patterns that truly belong to that specific class are still in majority in that set. Now the problem of robust learning under label noise can be stated as follows: We want to learn a classifier for the distribution $\mathcal{D}$ but given training data drawn from $\mathcal{D}_{\eta}$.

We denote by  $f(\cdot;\theta)$ a classifier function parameterized by $\theta$. We assume that the neural network classifiers that we use have softmax output layer. Hence, while  the training set labels, $y_i$, are one-hot vectors, we will have  $ f(x;\theta) \in \Delta^{K-1}$, where $ \Delta^{K-1} \subset [0,1]^K $ is the probability simplex. We denote by $\mathcal{L}(f(x;\theta),y)$ the loss function used for the classifier training which in our case is the CCE loss. 

\subsection{Adaptive Curriculum through Batch Statistics}
\label{sec:batch_statistics}

General curriculum learning can be viewed as minimization of a weighted loss \cite{spl,mentornet} 
\begin{equation}
	\begin{split}
		\min_{\theta, {\bf w} \in [0, 1]^m} \mathcal{L}_{\mbox{wtd}}(\theta, {\bf w}) & = \sum_{i=1}^m w_{i}\mathcal{L}(f(x_{i};\theta),y_{i}) \\
		& + G({\bf w}) + \beta ||\theta||^2
	\end{split}
\end{equation}
where $G({\bf w})$ represents the curriculum. Since one normally employs SGD for learning, we will take $m$ here to be the size of a mini-batch. One simple choice for the curriculum is \cite{spl}: $ G({\bf w}) = -\lambda ||{\bf w}||_1, \; \lambda > 0$. Putting this in the above, omitting the regularization term and taking $l_i = \mathcal{L}(f(x_{i};\theta),y_{i}) $, the optimization problem becomes
\begin{gather}
	\min_{\theta, {\bf w} \in [0, 1]^m} \mathcal{L}_{\mbox{wtd}}(\theta, {\bf w}) = \sum_{i=1}^m \left(w_{i} l_i - \lambda w_i\right) \\
	 = \sum_{i=1}^m \left(w_{i} l_i + (1-w_i) \lambda \right)  - m \lambda
	\label{eq:opt1}
\end{gather}
Under the usual assumption that loss function is non-negative, for the above problem, the optimal ${\bf w}$ for any fixed $\theta$ is: $w_i = 1$ if $l_i < \lambda$ and $w_i=0$ otherwise. 
%If we want an adaptive curriculum, we want $\lambda$ to be dynamically adjusted based on the current state of learning. 
We first consider a modification where we make $\lambda$ depend on the class label. The optimization problem becomes
\begin{gather}
	\min_{\theta, {\bf w} \in [0, 1]^m} \mathcal{L}_{\mbox{wtd}}(\theta, {\bf w}) = \sum_{i=1}^m \left(w_{i} l_i - \lambda(y_i) w_i\right) \\
	% & = & \sum_{j=1}^K \sum_{\substack{i=1 \\ i: y_i = e_j}}^{m} \left(w_{i} l_i - \lambda_j w_i\right) \\
	 = \sum_{j=1}^K \sum_{\substack{i=1 \\ i: y_i = e_j}}^{m} \left(w_{i} l_i + (1-w_i) \lambda_j \right)  - \sum_{j=1}^K \sum_{\substack{i=1 \\ i: y_i = e_j}}^{m} \lambda_j
	\label{eq:opt2}	
\end{gather}
where $\lambda_j = \lambda(e_j)$. As is easy to see, the optimal $w_i$ (for any fixed $\theta$) are still given by the same relation: for an $i$ with $y_i=e_j$, $w_i = 1$ when $l_i < \lambda_j$. Note that this relation for optimal $w_i$ is true even if we make $\lambda_j$ a function of $\theta$ and of all $x_i$ with $y_i = e_j$. Thus we can have a truly dynamically adaptive curriculum by making these $\lambda_j$ depend on all $x_i$ of that class in the mini-batch and the current $\theta$.

The above is an interesting insight: in the Self-Paced Learning formulation \cite{spl}, the nature of the final solution is same even if we make the $ \lambda $ parameter a function of the class-labels and also other feature vectors corresponding to that class. This gives rise to class-label-dependent thresholds on loss values. To the best of our knowledge, this direction of curriculum learning has not been explored. The next question is how should we decide or evolve these $\lambda_j$. As we mentioned earlier, we want these to be determined by the statistics of loss values in the mini-batch. 

Consider those $i$ for which $y_i=e_j$.  We would be setting $w_i=1$ and hence use this example to update $\theta$ in this minibatch if this $l_i < \lambda_j$. We want $\lambda_j$ to be fixed based on the observed loss values of this mini-batch. Since there is sufficient empirical evidence that we tend to learn from the clean samples before overfitting to the noisy ones, some quantile or similar statistic of the set of observed loss values in the mini-batch (among patterns labelled with a specific class)  would be a good choice for $\lambda_j$. 
%In this paper we experiment with two choices for this statistic: the median and the mean plus one standard deviation. Both result in similar performance

Since we are using CCE loss, we have $l_i = -\ln\left(f_j(x_{i};\theta)\right)$ and as the network has softmax output layer, $f_j(x_{i};\theta)$ is the posterior probability of class-$j$ under current $\theta$ for $x_i$. Since the loss and this posterior probability are inversely related, our criterion for selection of an example could be that the assigned posterior probability is above a threshold which is some statistic of the observed posterior probabilities in the mini-batch. In this paper we take the statistic to be mean plus one standard deviation.

In other words, in any mini-batch, we set the weights for samples as 
\begin{equation}
	w_{i}=\begin{cases}
		1 & \text{ if }f_{y_{i}}(\x_{i};\theta)\geq \lambda_{y_{i}} = \mu_{y_{i}}+ \kappa * \sigma_{y_{i}}\\
		0 & \text{ else}
	\end{cases}
	\label{eq:thresh}
\end{equation}
%\vskip 0.2cm
where $ \mu_{y_{i}} = \frac{1}{|\mathcal{S}_{y_i}|}\sum_{s\in\mathcal{S}_{y_i}}f_{y_{i}}(\x_{s};\theta) $ and $\sigma_{y_i}^2 = \frac{1}{|\mathcal{S}_{y_{i}}|}\sum_{s\in\mathcal{S}_{y_{i}}} (f_{y_{i}}(\x_{s};\theta) - \mu_{y_{i}})^2$ indicate the sample mean and sample variance of the class posterior probabilities for samples having class label $ y_{i} $. \textbf{[\underline{Note}:} $ \mathcal{S}_{y_{i}} = \{k\in[m]\;|\;y_{k}=y_{i}\}$ where $ m $ is the size of mini-batch\textbf{]}. We use $ \kappa = 1 $ in this paper but we empirically observe that as long as samples from the `top quantiles' are chosen (i.e. $ \kappa > 0 $), we get good and similar robustness against label noise across different $ \kappa $. See Table \ref{table:thresh-abl-bare} in Supplementary for an ablation study.

Figures \ref{fig:mnist-cls-thresh-sym}--\ref{fig:cifar-cls-thresh-cc} (in Supplementary) show that the threshold value (RHS of Equation \ref{eq:thresh} with $ \kappa = 1 $) varies across different mini-batches for a given class or epoch. This varying nature of statistics of the loss values in a mini-batch further justifies the rationale for our method of choosing an adaptive threshold.

%Algorithm \ref{alg:bare} summarizes the procedure.

\subsection*{Algorithm Implementation}

Algorithm \ref{alg:bare} outlines the proposed method. Keeping in mind that neural networks are trained in a mini-batch manner, Algorithm \ref{alg:bare} consists of three parts: i.) computing sample selection thresholds, $ \lambda_{y_{\x}} $, for a given mini-batch of data (Step 8-13), ii.) sample selection based on these thresholds (Steps 15-19) as per Equation \ref{eq:thresh}, and iii.) network parameter updation using these selected samples (Step 20).

\begin{algorithm}[ht]
	\caption{BAtch REweighting (BARE) Algorithm}
	\label{alg:bare}
	\begin{algorithmic}[1]
		\STATE {\bfseries Input:} noisy dataset $\mathcal{D}_{\eta}$, $\#$ of classes $K$, $\#$ of epochs $T_{max}$, learning rate $\alpha$, mini-batch size $|\mathcal{M}|$
		\STATE {\bfseries Initialize:} Network parameters, $\theta_{0} $, for classifier $ f(\cdot; \theta) $
		\vspace*{-0.4cm}
		\FOR{$t=0$ {\bfseries to} $T_{max} - 1$}
		\STATE Shuffle the training dataset $\mathcal{D}_{\eta}$
		\FOR{$i=1$ to $|\mathcal{D}_{\eta}|/|\mathcal{M}|$}
		\STATE Draw a mini-batch $\mathcal{M}$ from $\mathcal{D}_{\eta}$
		\STATE $ m = |\mathcal{M}| $ \quad // mini-batch size
		\FOR{$p=1$ to $ K $}
		\STATE $ \mathcal{S}_p = \{k\in[m]\;|\;y_{k}=e_{p}\} $ \\ {\small // collect indices of samples with class-$ p $}
		\STATE $ \mu_p = \frac{1}{|\mathcal{S}_{p}|}\sum_{s\in\mathcal{S}_{p}}f_{p}(\x_{s};\theta_{t}) $ \\ {\small // mean posterior prob. for samples with class-$ p $}
		\STATE {\footnotesize $ \sigma_{p}^2 = \frac{1}{|\mathcal{S}_{p}|}\sum_{s\in\mathcal{S}_{p}} (f_{p}(\x_{s};\theta_{t}) - \mu_p)^2 $ \\ // variance in posterior prob. for samples with class-$ p $}
		\STATE $\lambda_p$ $\leftarrow \mu_{p} + \sigma_{p}$ \quad {\small // sample selection threshold for class-$ p $ as per Equation \ref{eq:thresh}}
		\ENDFOR
		\STATE $\mathcal{R} \leftarrow \phi$ \quad // selected samples in $\mathcal{M}$
		\FOR{\textbf{each} $\x \in \mathcal{M}$}
		\IF{$ f_{y_{\x}}(\x;\theta_{t})\geq \lambda_{y_{\x}} $}
		\STATE $\mathcal{R} \leftarrow \mathcal{R} \cup (\x, y_{\x})$ \\ // Select sample as per Equation \ref{eq:thresh}
		\ENDIF
		\ENDFOR
		%		\STATE $\mathbb{R} \leftarrow \mathbb{R} \cup \mathcal{R}$
		\STATE $\mathbf{\theta}_{t+1} = \mathbf{\theta}_{t} - \alpha\nabla\Big(\frac{1}{|\mathcal{R}|}\sum_{(\x,y_{\x})\in\mathcal{R}}\mathcal{L}(\x,y_{\x};\theta_{t})\Big)$ \quad // parameter updates
		%		\STATE Update $\mathbf{w}^{t} = \underset{\mathbf{w} \in \{0, 1\}^{m}}{\text{arg min}}\sum_{i=1}^{m}\Big[w_{i}\mathcal{L}(f(x_{i};\mathbf{\theta}^{t-1}),y_{i})+(1-w_{i})K(\{x_{i}\}_{i=1}^{m},\theta^{t-1})\Big] \qquad\text{// sample selection}$
		\ENDFOR
		\ENDFOR
		\STATE {\bfseries Output:} $\theta_{t}$ %, \mathbb{R}$
	\end{algorithmic}
\end{algorithm}

\section{Experiments on Noisy Dataset}
\label{sec:exp}

%\subsection{Experimental Setup}
%\label{sec:exp-setup}

\noindent\textbf{\underline{Dataset}:} We demonstrate the effectiveness of the proposed algorithm on two benchmark image datasets: MNIST and CIFAR10. These data sets are used to benchmark almost all algorithms for robust learning under label noise and we briefly describe the data sets. MNIST contains 60,000 training images and 10,000 test images (of size $28 \times 28$) with 10 classes. CIFAR-10 contains 50,000 training images and 10,000 test images (of size $32 \times 32$) with 10 classes. We test the algorithms on two types of label noise: symmetric and  class-conditional label noise. In symmetric label noise, each label is randomly flipped to any of the remaining classes with equal probability, whereas for class-conditional noise, label flipping is done in a set of similar classes. For the simulations here, for MNIST, the following flipping is done: $1 \leftarrow 7$, $ 2 \rightarrow 7 $, $ 3 \rightarrow 8 $, and $ 5 \leftrightarrow 6 $. Similarly, for CIFAR10, the following flipping is done: TRUCK $ \rightarrow $ AUTOMOBILE, BIRD $ \rightarrow $ AIRPLANE, DEER $ \rightarrow $ HORSE, CAT $ \leftrightarrow $ DOG. We use this type of noise because it is arguably a more realistic scenario and also because it is the type of noise, in addition to symmetric noise, that other algorithms for learning under label noise have used. We also provide results with an arbitrary noise rate matrix (see Supplementary).  For all the datasets, 80\% of the training set is used for training and, from the remaining 20\% data, we sample 1000 images that constitute the validation set.

We also experiment with the Clothing-1M dataset \cite{clothing1M} which is a large-scale dataset obtained by scrapping off the web for different images related to clothing. It contains noise that can be characterized as somewhat close to feature-dependent noise, the most generic kind of label noise. An estimated 40\% images have noisy labels. The training dataset contains 1 million images and the number of classes are 14. There are additional training, validation, and test sets of 50k, 14k, and 10k images respectively with clean labels. Since there's a class imbalance, following similar procedure as in existing baselines, we use 260k images from the original noisy training set for training while ensuring equal number of images per class in the set and test set of 10k images for performance evaluation.

\noindent\textbf{\underline{Data Augmentations}:} No data augmentation is used for MNIST. Random croppings with padding of 4, and random horizontal flips are used for CIFAR-10. For Clothing-1M. we do random cropping while ensuring image size is fixed.

\noindent\textbf{\underline{Baselines}:} We compare the proposed algorithm with the following algorithms from literature: 1.) \textbf{Co-Teaching (CoT)} \cite{coteaching} which involves cross-training of two similar networks by selecting a fraction (dependent on noise rates) of low loss valued samples; 2.) \textbf{Co-Teaching+ (CoT+)} \cite{coteaching+} which improves upon CoT with the difference being sample selection only from the subset upon which the two networks' predictions disagree; 3.) \textbf{Meta-Ren (MR)} \cite{meta-ren}, which involves meta-learning of sample weights on-the-fly by comparing gradients for clean and noisy data; 4.) \textbf{Meta-Net (MN) \cite{meta-net}}, which improves upon MR by explicitly learning sample weights via a separate neural network; \textbf{Curriculum Loss (CL)} \cite{curr-loss}, which involves a curriculum for sample selection based on (estimated) noise rates; and 6.) \textbf{Standard (CCE)} which is the usual training through empirical risk minimization with cross-entropy loss (using the data with noisy labels).

Among these baselines, CoT, CoT+, and CL are sample selection algorithms that require knowledge of  noise rates. The algorithms CoT+ and CL need a few initial iterations without any sample selection as  a warm-up period; we used 5 epochs and 10 epochs as warm up period during training for MNIST and CIFAR-10 respectively. MR and MN assume access to a small set of clean validation data. Because of this, and for a fair comparison among all the baselines, a clean validation set of 1000 samples is used in case of MR and MN, and the same set of samples but with the noisy labels is used for the rest of the algorithms including the proposed one.  % For all algorithms we compare test accuracies on a separate test set with clean labels.

\noindent\textbf{\underline{Network architectures \& Optimizers}:} While most algorithms for learning under label noise use MNIST and CIFAR10 data, different algorithms use different network architectures. Hence, for a fairer comparison, we have decided to use small networks that give state of art performance on clean data and investigate the robustness we get by using our algorithm on these networks. Please refer the supplementary material for details about the network architectures and optimization routines.

\noindent\textbf{\underline{Performance Metrics}:} For all algorithms we compare \textbf{test accuracies} on a separate test set with clean labels. The main idea in all sample selection schemes is to identify noisy labels. Hence, in addition to test accuracies, we also compare \textbf{precision} ($\#$ \textit{clean labels selected / $\#$ of selected labels}) and \textbf{recall} ($\#$ \textit{clean labels selected / $\#$ of clean labels in the data}) in identifying noisy labels. % We use 1.) test accuracy ($\#$ \textit{correct predictions / size of test dataset}), 2.) label precision ($\#$ \textit{clean labels selected / $\#$ of selected labels}), and 3.) label recall ($\#$ \textit{labels not selected / $\#$ of clean labels}) as the metrics for comparison of various baselines.

\begin{figure*}[ht!]
	%	\subfloat{\label{fig:mnist-sym-02-acc}\includegraphics[scale=0.4]{plots/sample_rewgt/mnist/sym/sample_rewgt_mnist_nr_02_nt_sym.jpg}}
	\hspace*{-0.45cm} 
\subfloat[]{\label{fig:mnist-sym-05-acc}\includegraphics[scale=0.03]{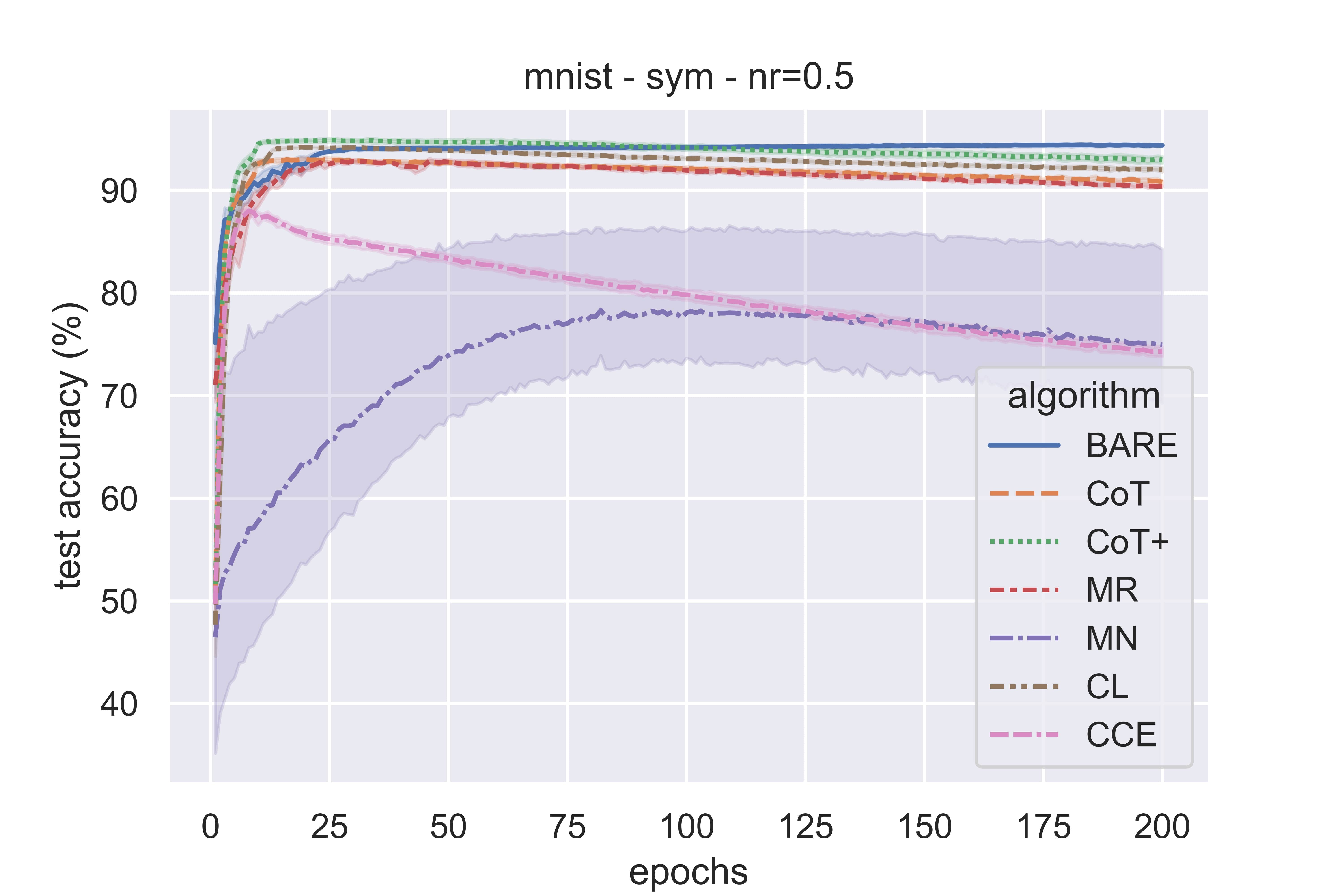}}
	\subfloat[]{\label{fig:mnist-sym-07-acc}\includegraphics[scale=0.03]{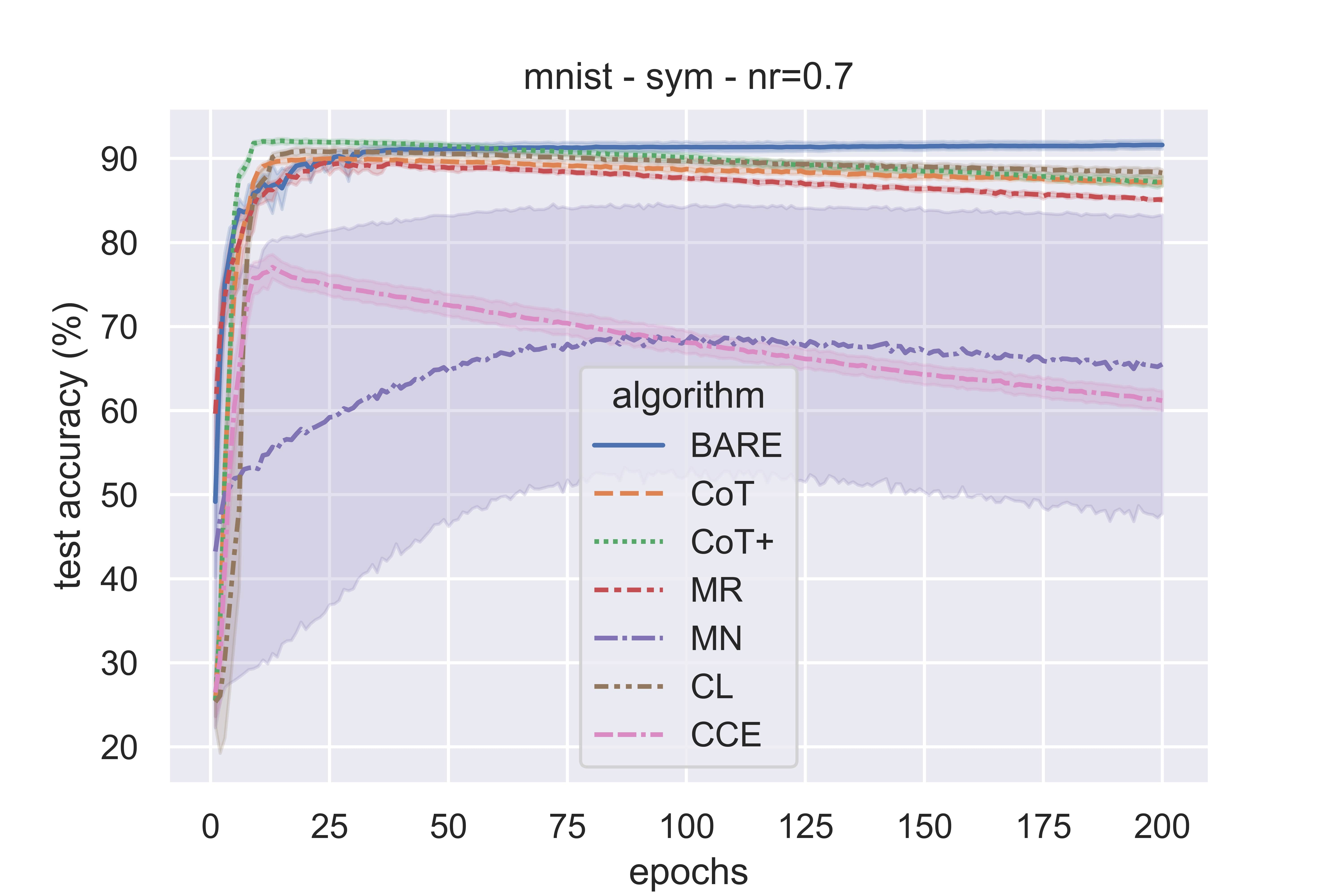}}
	\subfloat[]{\label{fig:mnist-cc-045-acc}\includegraphics[scale=0.03]{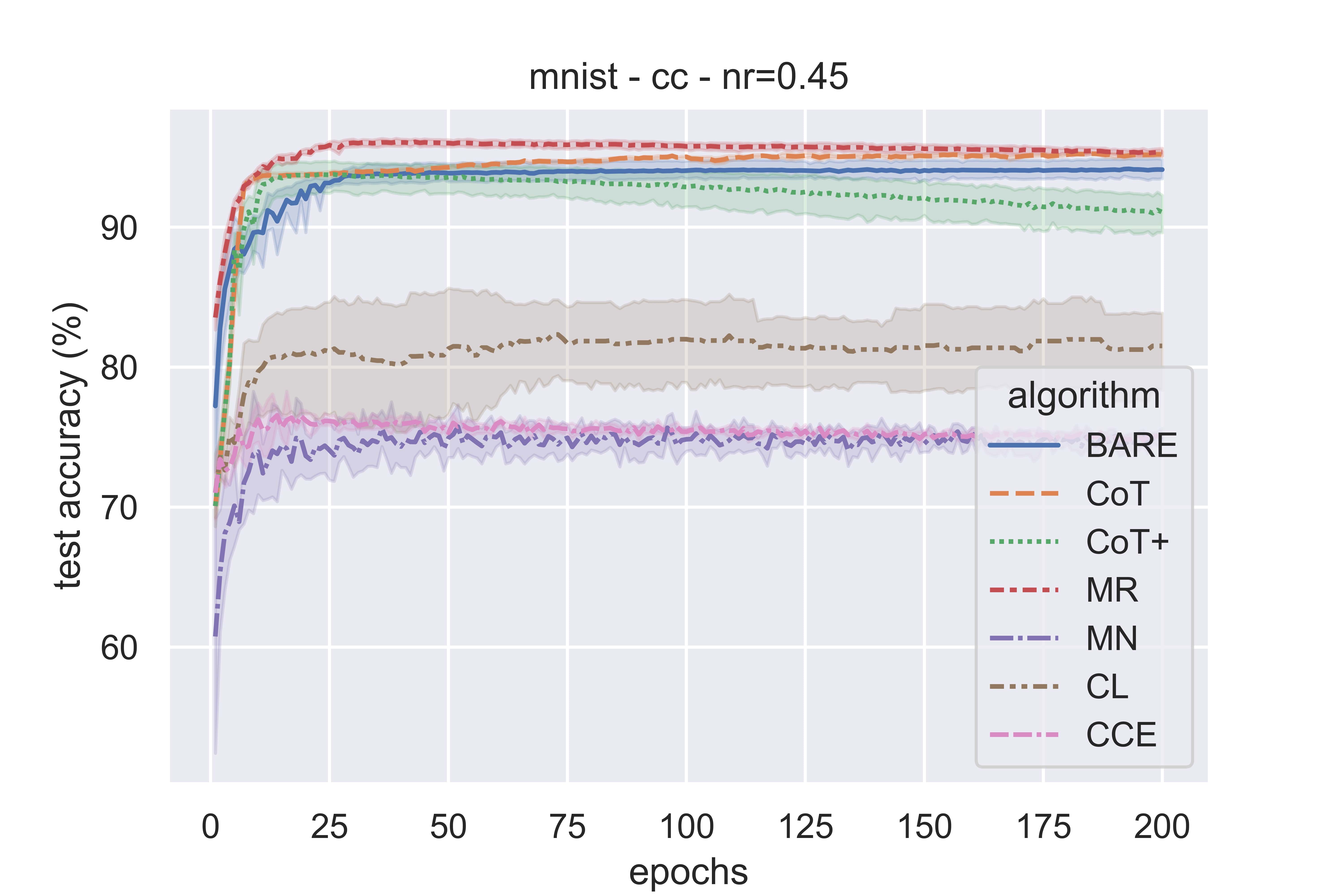}}
	\caption{Test Accuracies - MNIST - Symmetric (\subref{fig:mnist-sym-05-acc} \& \subref{fig:mnist-sym-07-acc}) \& Class-conditional (\subref{fig:mnist-cc-045-acc}) Label Noise}
	\label{fig:mnist}	
\end{figure*}

\begin{figure*}[ht!]
	\hspace*{-0.45cm}
	\subfloat[]{\label{fig:cifar10-sym-03-acc}\includegraphics[scale=0.03]{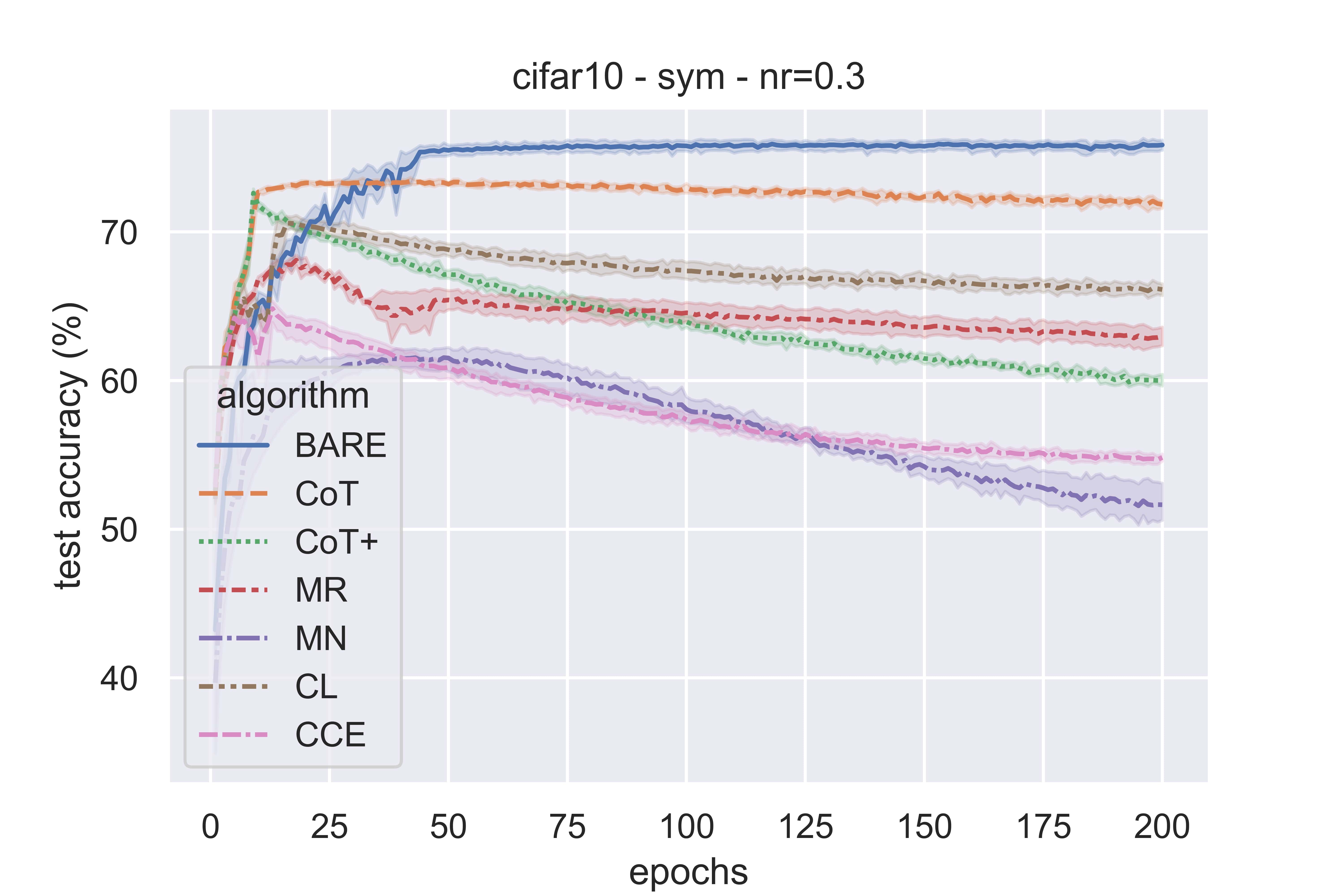}}
	\subfloat[]{\label{fig:cifar10-sym-07-acc}\includegraphics[scale=0.03]{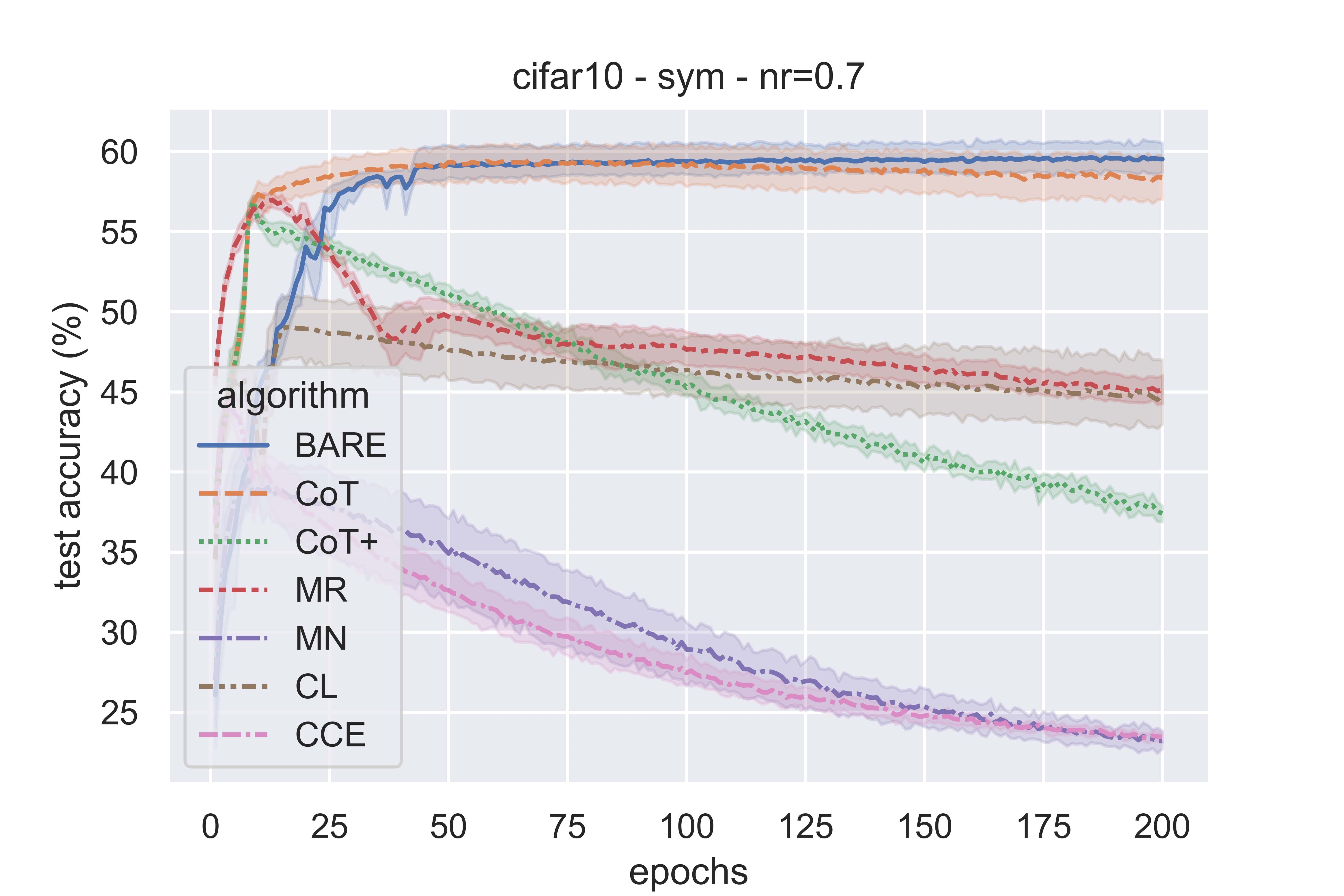}}
	\subfloat[]{\label{fig:cifar10-cc-04-acc}\includegraphics[scale=0.03]{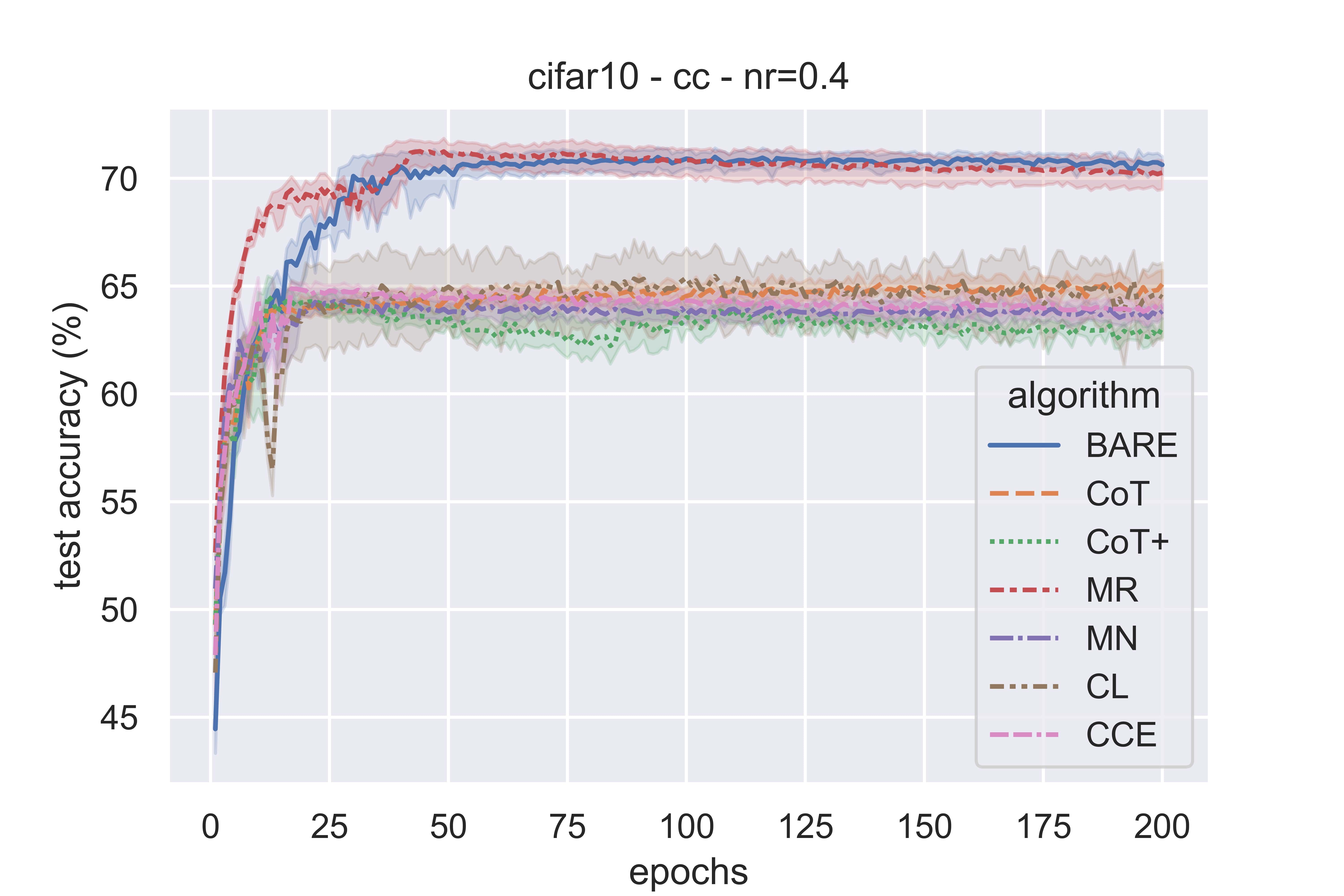}}
	\caption{Test Accuracies - CIFAR10 - Symmetric (\subref{fig:cifar10-sym-03-acc} \& \subref{fig:cifar10-sym-07-acc}) \& Class-conditional (\subref{fig:cifar10-cc-04-acc}) Label Noise}
	\label{fig:cifar10}
\end{figure*}

\begin{figure*}[h!]
	%	\subfloat{\label{fig:mnist-sym-02-lab-prec}\includegraphics[scale=0.4]{plots/sample_rewgt/mnist/sym/sample_rewgt_mnist_nr_02_nt_sym.jpg}}
	\hspace*{-0.45cm}
	\subfloat[]{\label{fig:mnist-sym-05-lab-prec}\includegraphics[scale=0.03]{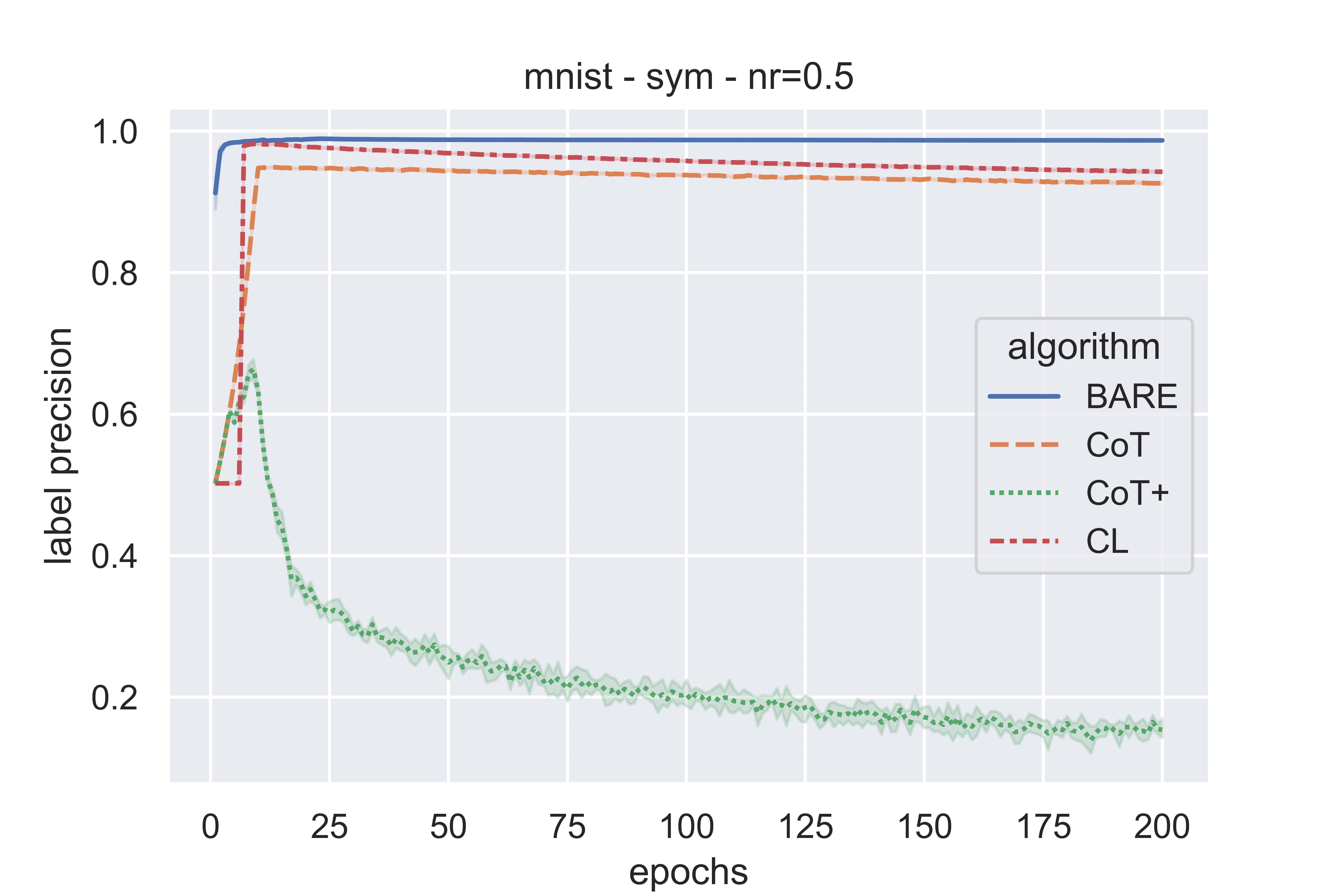}}
	\subfloat[]{\label{fig:mnist-sym-07-lab-prec}\includegraphics[scale=0.03]{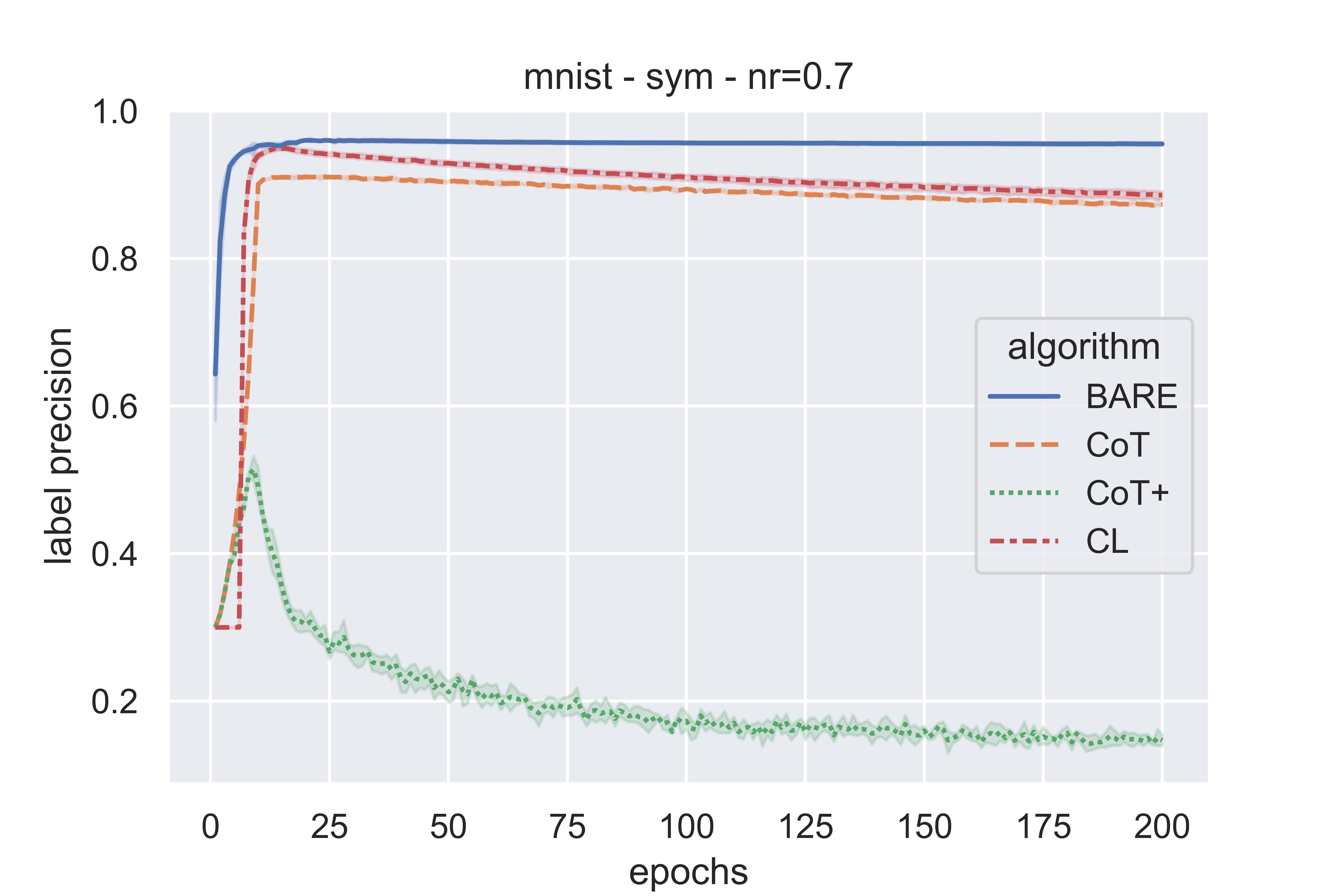}}
	\subfloat[]{\label{fig:mnist-cc-045-lab-prec}\includegraphics[scale=0.03]{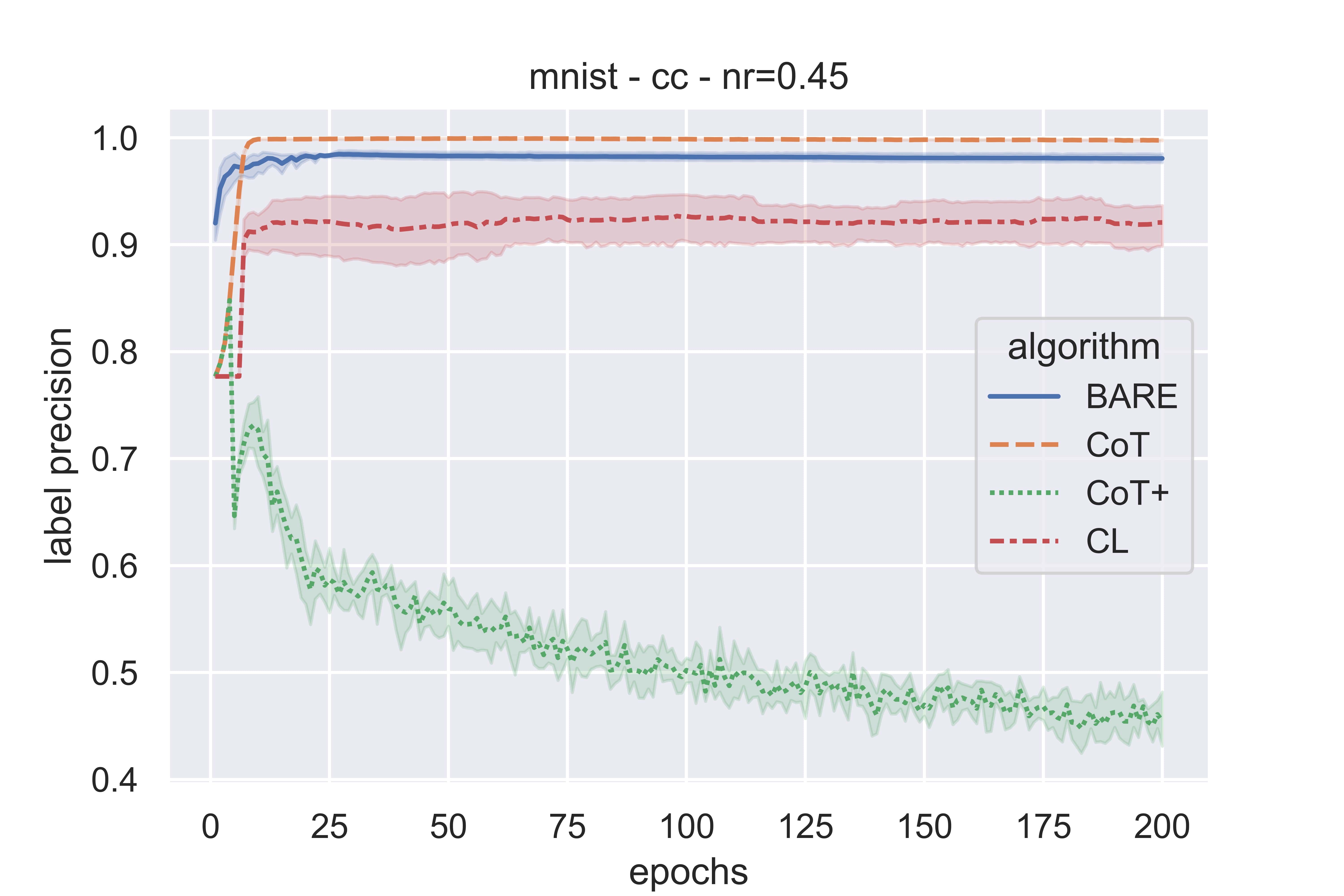}}
	\caption{Label Precision - MNIST - Symmetric (\subref{fig:mnist-sym-05-lab-prec} \& \subref{fig:mnist-sym-07-lab-prec}) \& Class-conditional (\subref{fig:mnist-cc-045-lab-prec}) Label Noise}
	\label{fig:mnist-lab-prec}
\end{figure*}

\vspace*{-0.1cm}
\subsection{Discussion of Results}

\vspace*{-0.1cm}
\textbf{Performance on MNIST.}  Figure \ref{fig:mnist} shows the evolution of test accuracy (with training epochs) under symmetric ($ \eta\in\{0.5,0.7\} $) and class conditional ($\eta = 0.45$) label noise for different algorithms.
%Figure \ref{fig:mnist-sym-05-acc} and \ref{fig:mnist-sym-07-acc} show the test accuracies under various noise rates for symmetric label noise while Figure \ref{fig:mnist-cc-045-acc} shows test accuracies for class-conditional label noise. 
We can see from the figure that the proposed algorithm outperforms the baselines for symmetric noise. For the case of class-conditional noise, the test accuracy of the proposed algorithm is marginally less than the best of the baselines, namely CoT and MR. 
%as good as or better than all baselines except CoT and MR. 
%One can note from Figure \ref{fig:mnist-lab-prec} that the proposed algorithm consistently selects higher fraction of clean samples for training compared the other algorithms. %However, for class-conditional label noise, it seems to be performing a bit poorly compared to CoT. \\

% We tabulate the final test accuracies of all algorithms in Tables \ref{table:mnist-sym-05-acc} -- \ref{table:mnist-cc-045-acc}. The best two results are in bold. These are accuracies achieved at the end of training. For CoT \cite{coteaching} and CoT+ \cite{coteaching+}, we show accuracies only of that network which performs the best out of the two that are trained.

\textbf{Performance on CIFAR-10.}  Figure \ref{fig:cifar10} shows the test accuracies of the various algorithms as the training progresses for both symmetric ($ \eta\in\{0.3, 0.7\} $) and class-conditional ($\eta = 0.4$) label noise. %The observations are quite similar to those seen for MNIST. 
%Figure \ref{fig:cifar10-sym-03-acc} and \ref{fig:cifar10-sym-07-acc} show the test accuracies under various noise rates for symmetric label noise whereas Figure \ref{fig:cifar10-cc-04-acc} shows test accuracies for class-conditional label noise ($\eta = 0.45$). 
We can see from the figure that the proposed algorithm outperforms the baseline schemes and its test accuracies are uniformly good for all types of label noise. It is to be noted that while test accuracies for our algorithm stay saturated after attaining maximum performance, the other algorithms' performance seems to deteriorate as can be seen in the form of accuracy dips towards the end of training. This suggests that our proposed algorithm doesn't let the network overfit even after long durations of training unlike the case with other algorithms. 

All the algorithms, except the proposed one, have hyperparameters (in the sample selection/weighting method) and the accuracies reported here are for the best possible hyperparameter values obtained through tuning. The MR and MN algorithms are particularly sensitive to hyperparameter values in the meta learning algorithm. In contrast, BARE has no hyperparameters for the sample selection and hence no such tuning is involved. It may be noted for the test accuracies on MNIST and CIFAR-10 that sometimes the standard deviation in the accuracy for MN is high. As we mentioned earlier, we noticed that MN is very sensitive to the tuning of hyper parameters. While we tried our best to tune all the hyper parameters, may be the final ones we found for these cases are still not the best and that is why the standard deviation is high.

% Unlike in case of MNIST, for CIFAR10, the performance of the proposed algorithm is superior to the baselines for both the symmetric and class-conditional label noise. 

%Please refer to Appendix ?? for the tabulated statistics (mean and standard deviation) of algorithm performances for all the experiments discussed in this section.

\begin{figure*}[h!]
	\hspace*{-0.45cm}
	\subfloat[]{\label{fig:cifar10-sym-03-lab-prec}\includegraphics[scale=0.03]{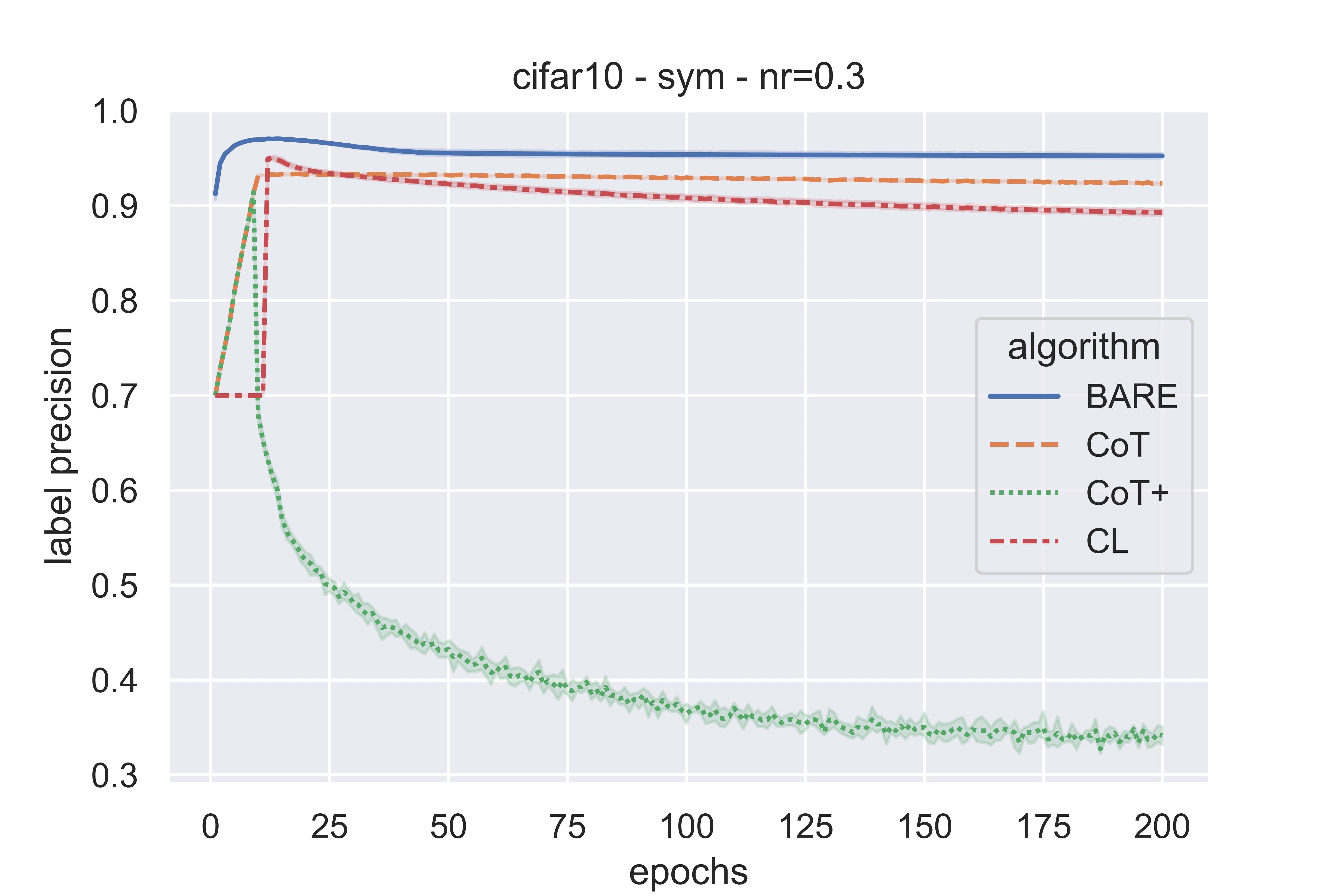}}
	\subfloat[]{\label{fig:cifar10-sym-07-lab-prec}\includegraphics[scale=0.03]{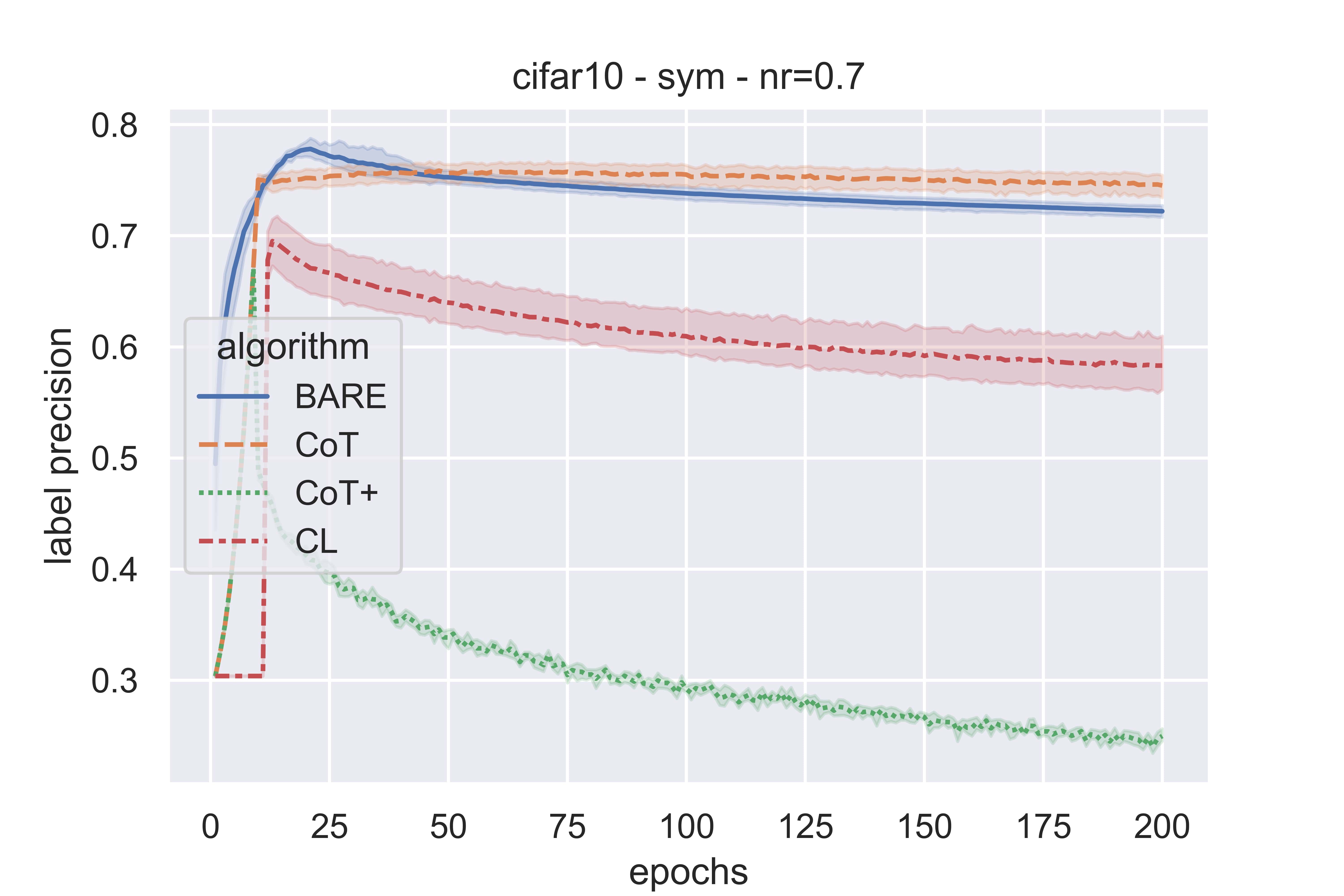}}
	\subfloat[]{\label{fig:cifar10-cc-04-lab-prec}\includegraphics[scale=0.03]{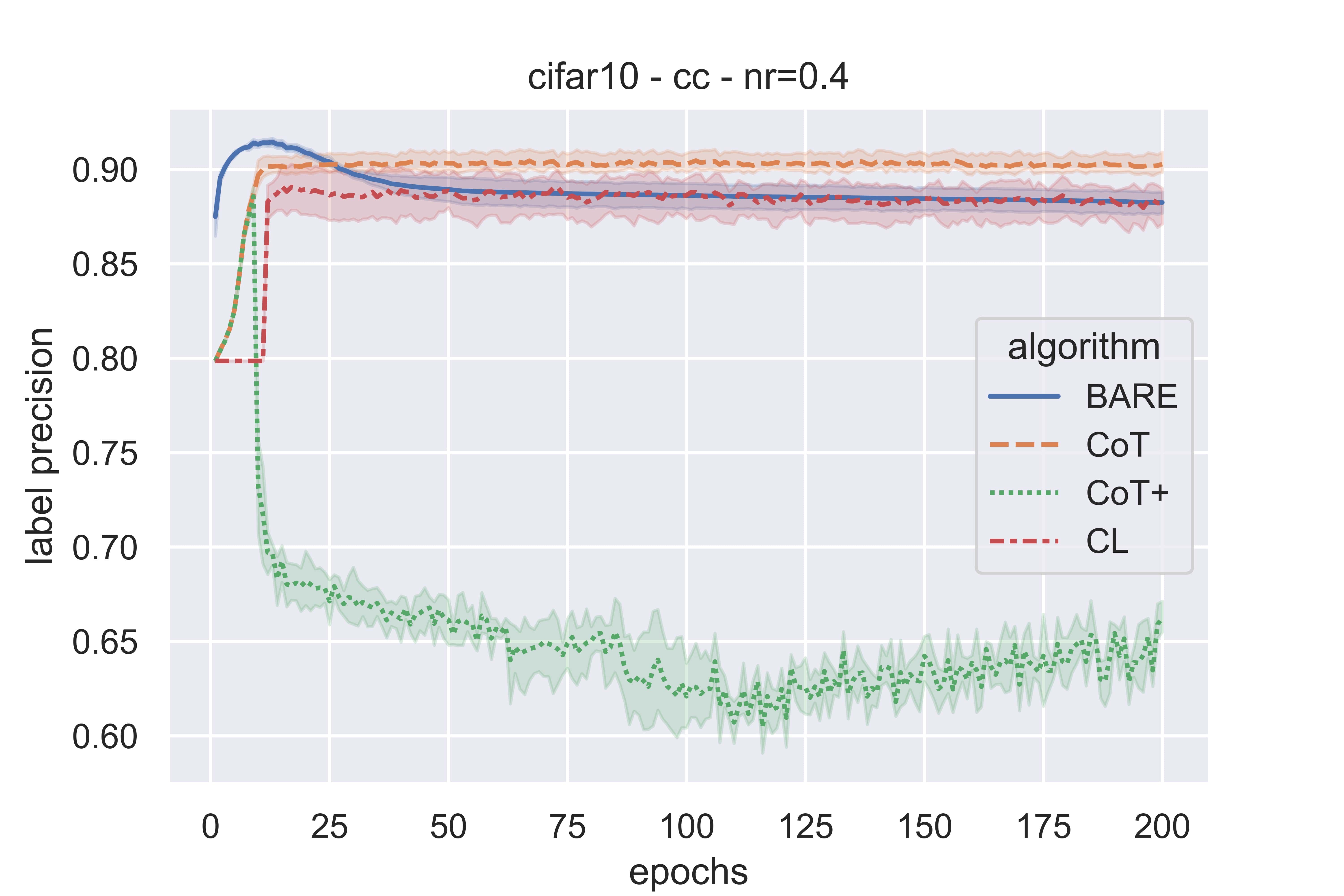}}
	\caption{Label Precision - CIFAR10 - Symmetric (\subref{fig:cifar10-sym-03-lab-prec} \& \subref{fig:cifar10-sym-07-lab-prec}) \& Class-conditional (\subref{fig:cifar10-cc-04-lab-prec}) Label Noise}
	\label{fig:cifar10-lab-prec}
\end{figure*}

\begin{figure*}[h!]
	
	\hspace*{-0.45cm}
	\subfloat[]{\label{fig:mnist-sym-05-lr}\includegraphics[scale=0.03]{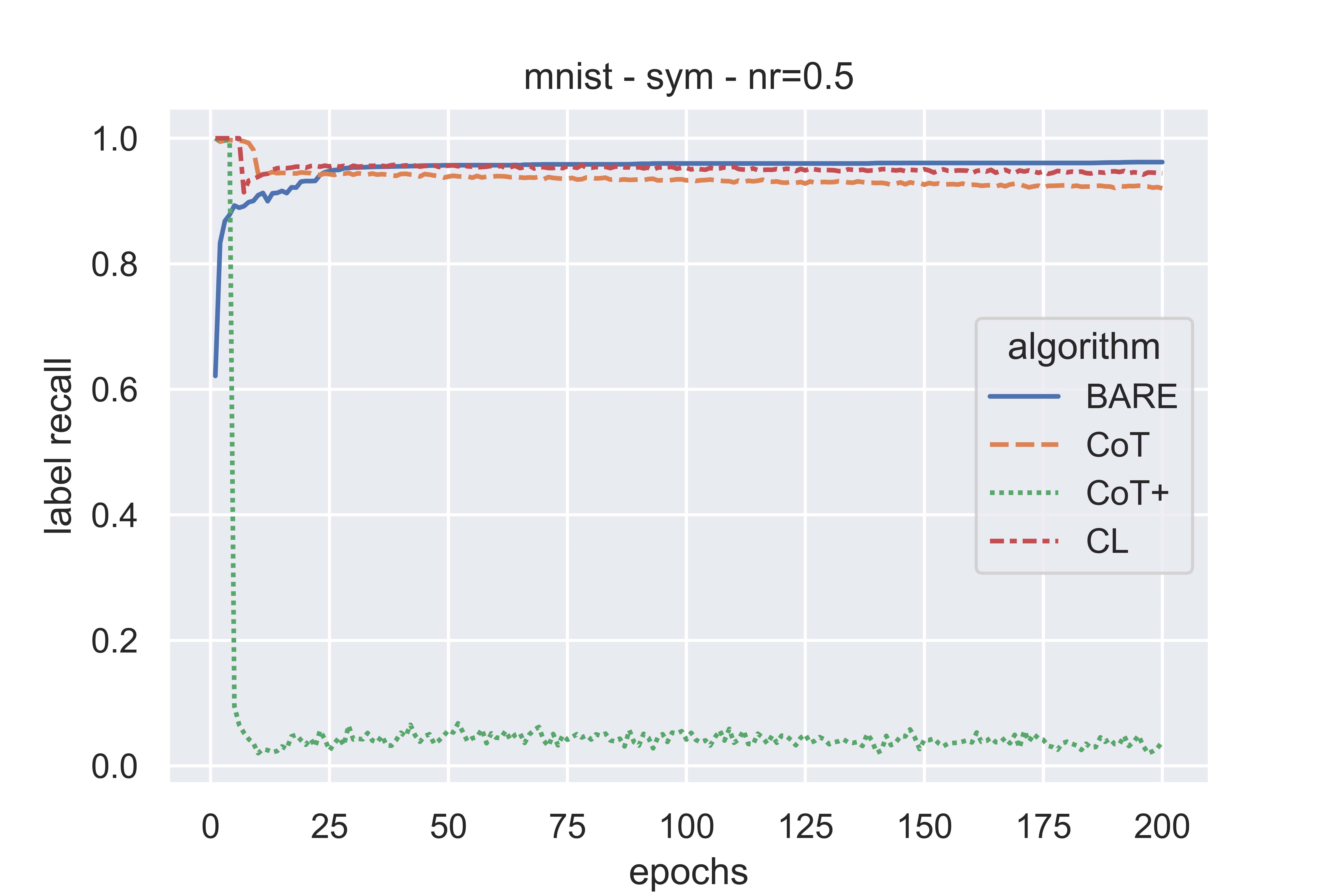}}
	\subfloat[]{\label{fig:cifar10-sym-07-lr}\includegraphics[scale=0.03]{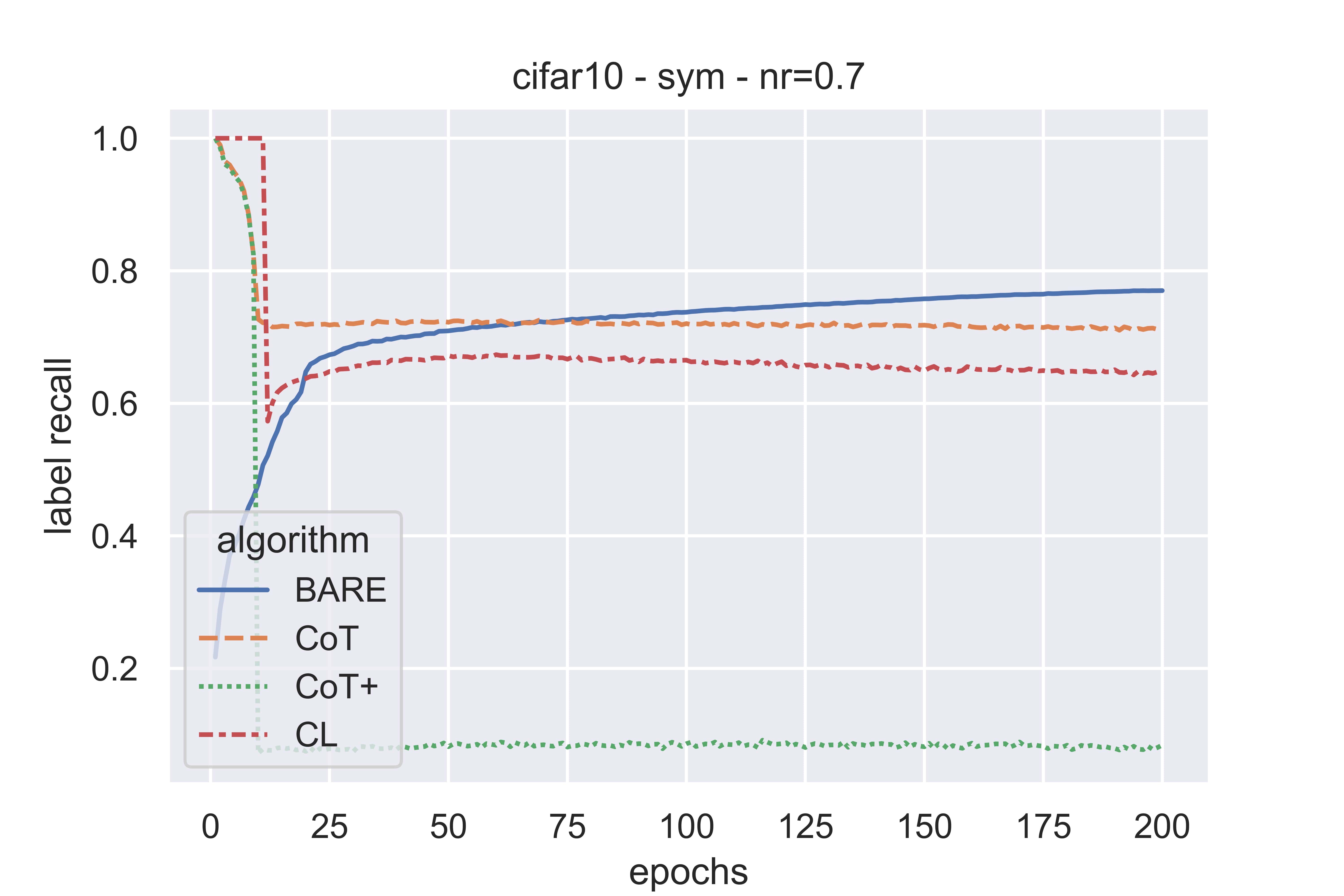}}
	\subfloat[]{\label{fig:cifar10-cc-045-lr}\includegraphics[scale=0.03]{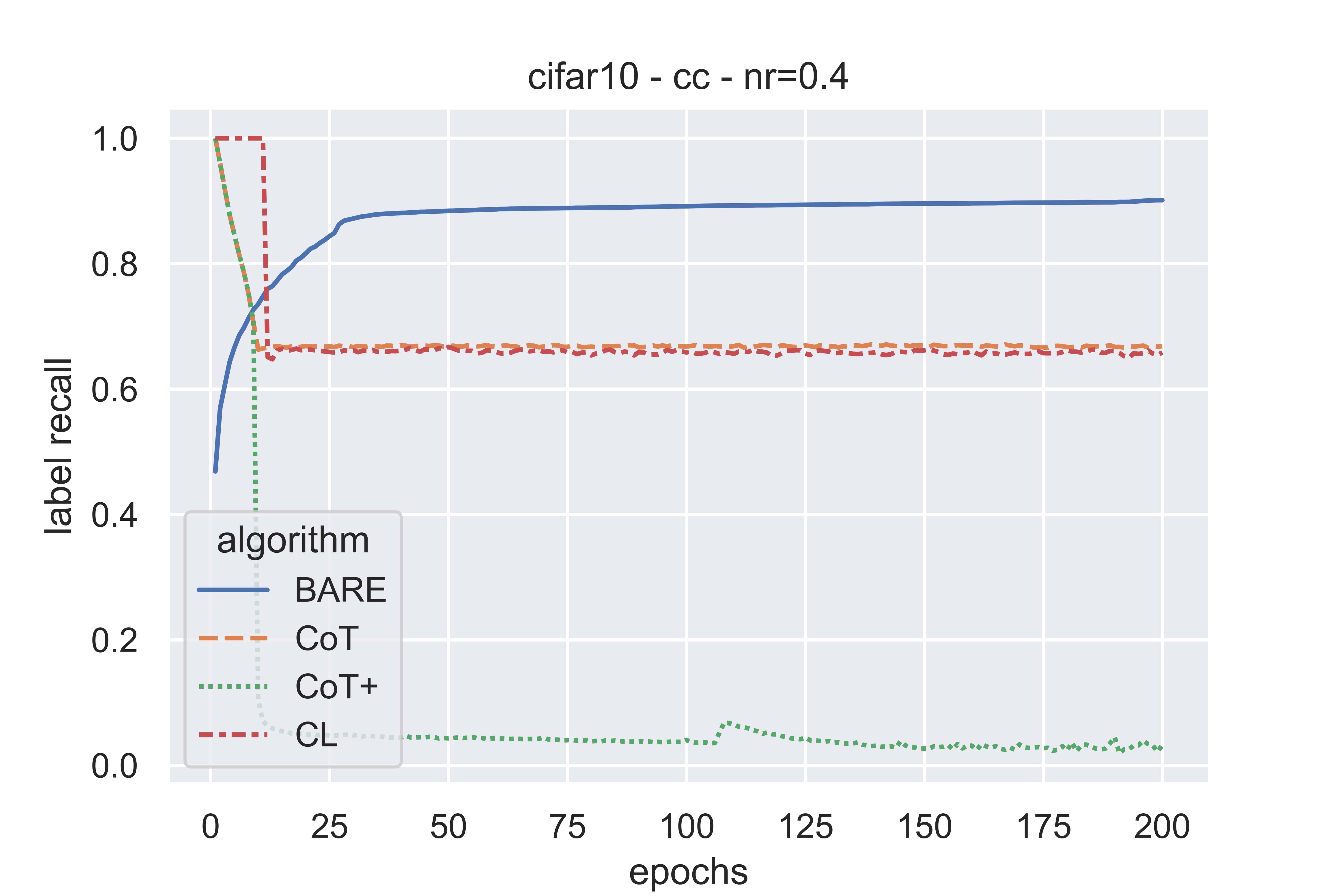}}
	\caption{Label Recall - Symmetric (\subref{fig:mnist-sym-05-lr} \& \subref{fig:cifar10-sym-07-lr}) \& Class-conditional (\subref{fig:cifar10-cc-045-lr}) Label Noise}
	\label{fig:mnist-cifar10-lr}	
\end{figure*}

\textbf{Performance on Clothing1M.} On this dataset, BARE achieved a test accuracy of 72.28\% against the accuracy of 68.8\% achieved by CCE. The accuracy achieved by BARE is better than that reported in the corresponding papers for all other baselines except for  C2D \cite{contrast-self-2022} \& DivideMix \cite{dividemix} which reported accuracy of 74.58\% \& 74.76\% resp. (The results are summarized in Table \ref{table:clothing-1m} in the Supplementary). These results show that even for datasets used in practice which have feature-dependent label noise, BARE performs better than all but two baselines. We note that the best performing baseline, DivideMix, requires about 2.4 times the computation time required for BARE. In addition to this, DivideMix requires tuning of 5 hyperparameters whereas no such tuning is required for BARE. The second best performing baseline, C2D, is also computationally expensive than BARE as it relies on self-supervised learning. % which makes it computationally expensive than BARE as well.

%\textbf{Performance on NEWS.} The results are summarized in Table \ref{table:news-sym}. Figure \ref{fig:news} shows the test accuracies as the training progresses for both symmetric and class-conditional label noise. \\

\begin{figure*}[h]
	\hspace*{-0.45cm}
	\subfloat[]{\label{fig:mnist-sent}\includegraphics[scale=0.03]{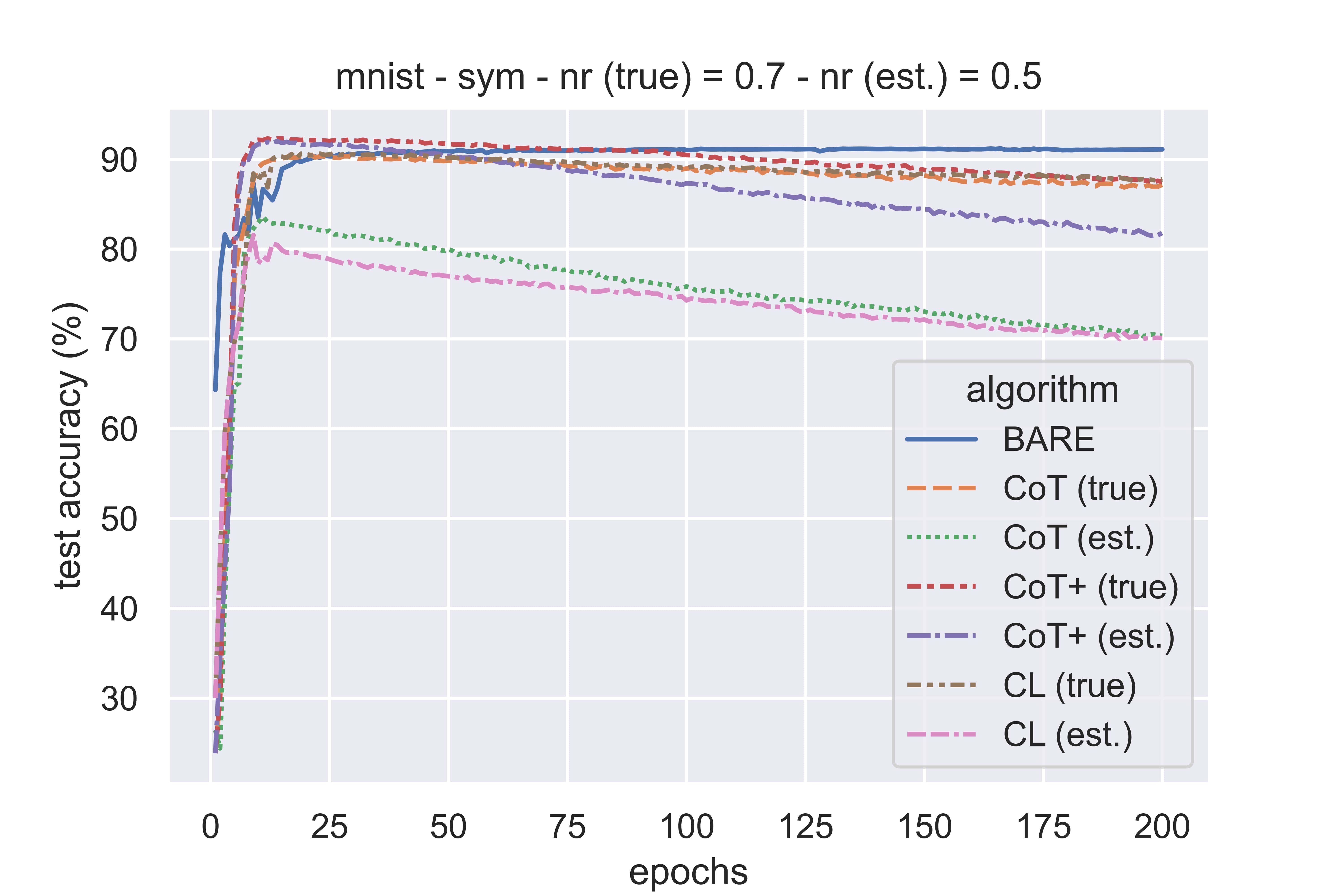}} \subfloat[]{\label{fig:cifar10-sent}\includegraphics[scale=0.03]{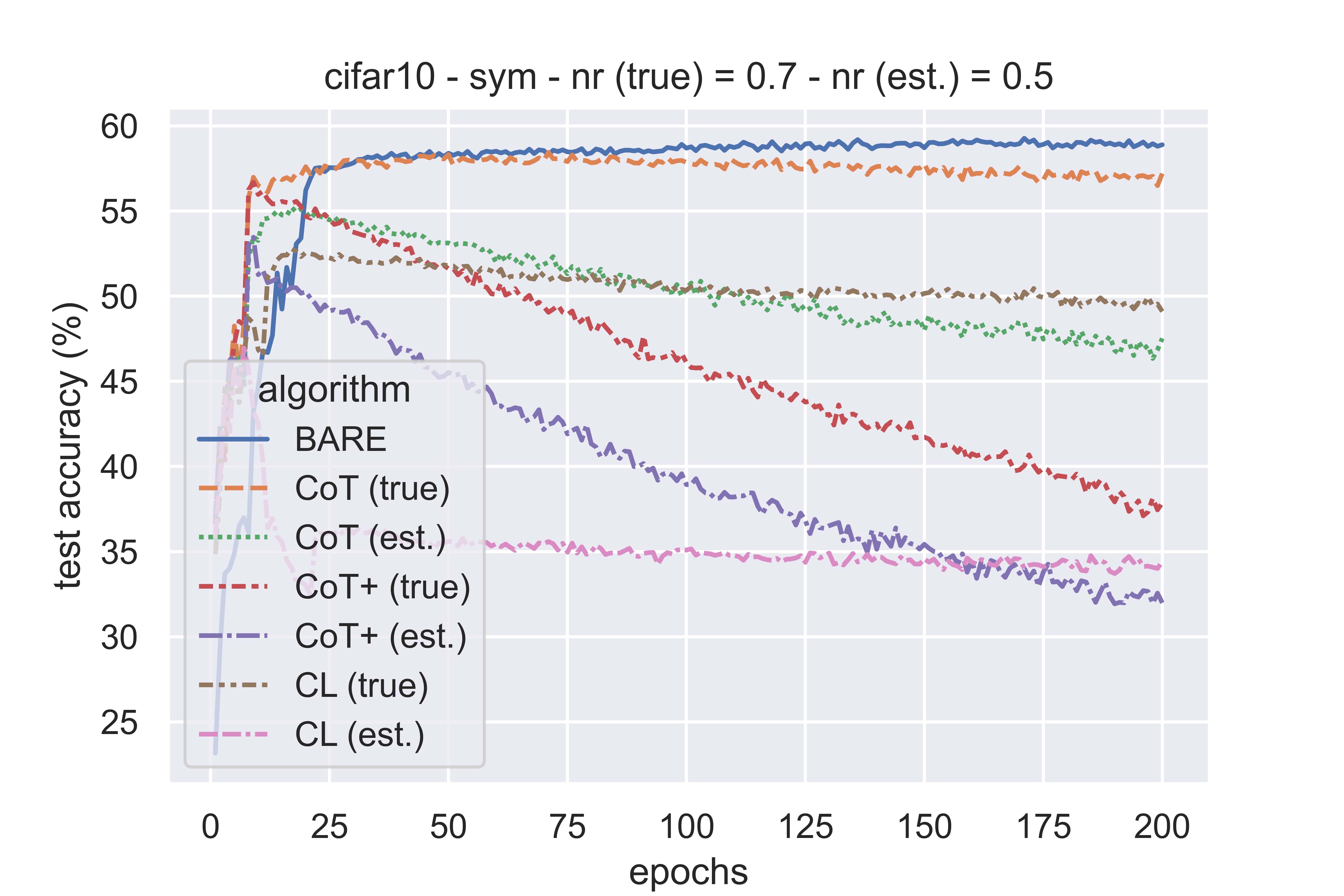}}
	\subfloat[]{\label{fig:cifar10-cc-frac}\includegraphics[scale=0.03]{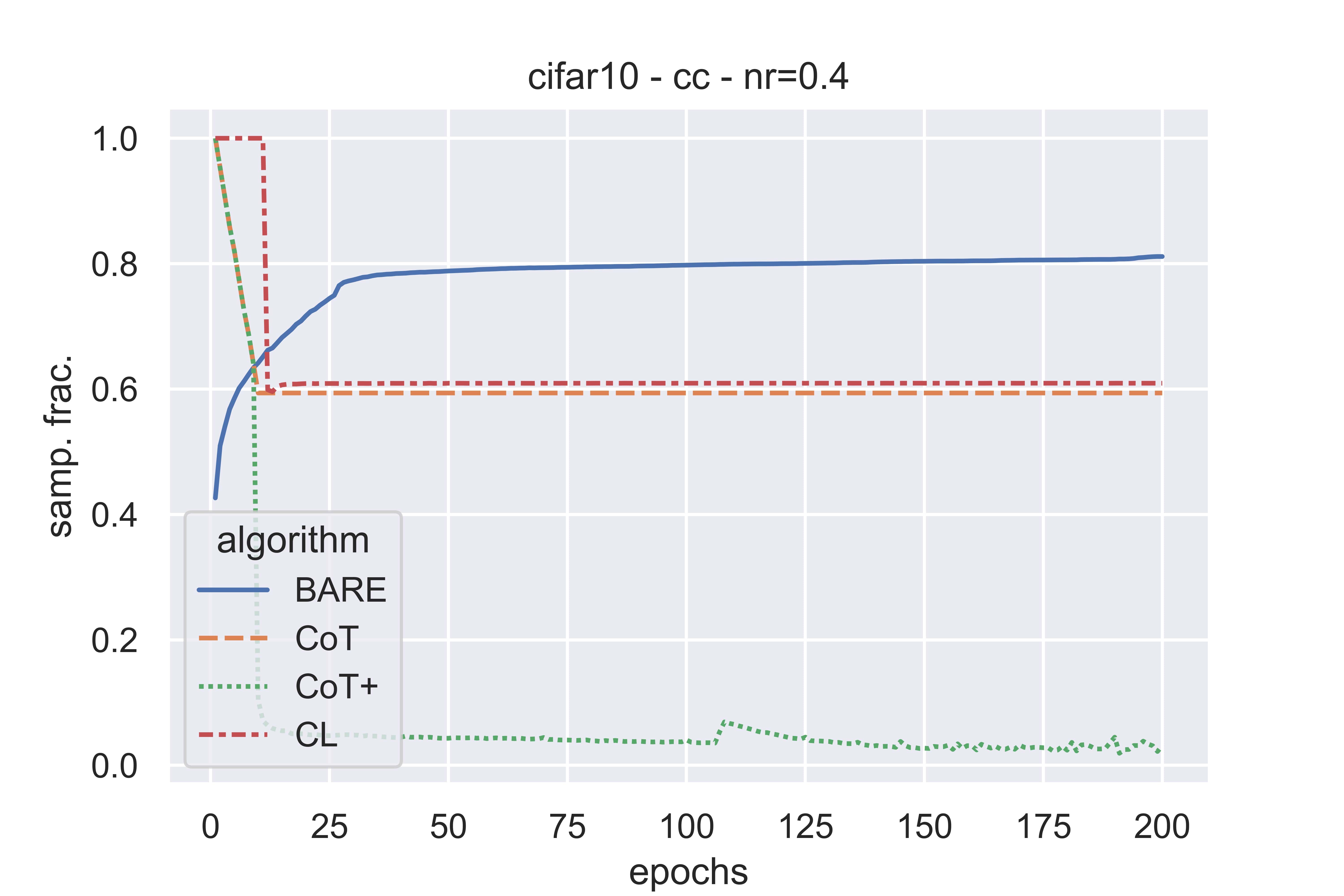}}
	
	%	\hspace*{-.5cm}\subfloat[]{\label{fig:mnist-sent-2}\includegraphics[scale=0.4]{./plots/sample_rewgt/mnist/sym/sample_rewgt_mnist_nr_05_nt_sym-samp-frac-fin.jpg}} \subfloat[]{\label{fig:cifar10-sent-2}\includegraphics[scale=0.4]{./plots/sample_rewgt/cifar10/sym/sample_rewgt_cifar10_nr_07_nt_sym-samp-frac-fin-correction.jpg}} 	

	\caption{(\subref{fig:mnist-sent} \& \subref{fig:cifar10-sent}): Test accuracies when estimated (symmetric) noise rate, $\eta=0.5$, and true noise rate, $\eta=0.7$, for MNIST \& CIFAR-10 resp.; (\subref{fig:cifar10-cc-frac}): sample fraction values for $\eta = 0.4$ (class-conditional noise) on CIFAR-10}
	
	%	\caption{(\subref{fig:mnist-sent-2}): Sample fraction values for $\eta = 0.5$ (symmetric noise) on MNIST, (\subref{fig:cifar10-sent-2}): Sample fraction values for $\eta = 0.7$ (symmetric noise) on CIFAR-10, (\subref{fig:cifar10-cc-frac}): sample fraction values for $\eta = 0.4$ (class-conditional noise) on CIFAR-10}
	\label{fig:mnist-cifar10-sent-samp-frac-2}	
\end{figure*}

%\begin{figure*}[h]
%		\centering
%		\subfloat[]{\label{fig:mnist-sent}\includegraphics[scale=0.5]{./plots/sample_rewgt/mnist/sym/sample_rewgt_mnist_nr_05_nt_sym-sent.jpg}} \subfloat[]{\label{fig:cifar10-sent}\includegraphics[scale=0.5]{./plots/sample_rewgt/cifar10/sym/sample_rewgt_cifar10_nr_05_nt_sym-sent.jpg}}
%
%	\caption{Test accuracies when estimated (symmetric) noise rate, $\eta=0.5$, and true noise rate, $\eta=0.7$, for (\subref{fig:mnist-sent}): MNIST \& (\subref{fig:cifar10-sent}): CIFAR-10}
%	\label{fig:mnist-cifar10-sent-samp-frac}
%\end{figure*}

\textbf{Efficacy of detecting clean samples.} Figure \ref{fig:mnist-lab-prec} and Figure \ref{fig:cifar10-lab-prec} show the label precision (across epochs) of the various algorithms on MNIST and CIFAR-10 respectively. 
One can see from these figures that BARE has comparable or better precision. Thus, compared to other sample selection algorithms, a somewhat higher fraction of examples selected for training by BARE have clean labels.

While \textit{test accuracies} and \textit{label precision} values do demonstrate the effectiveness of algorithms, it's also instructive to look at the label recall values. Label recall tells us how a sample selection algorithm performs when it comes to selecting reliable, clean samples. Figure \ref{fig:mnist-cifar10-lr} shows the label recall values for CoT, CoT+, CL, and BARE for MNIST (\ref{fig:mnist-sym-05-lr}) and CIFAR-10 (\ref{fig:cifar10-sym-07-lr} \& \ref{fig:cifar10-cc-045-lr}). It can be noted that BARE consistently achieves better recall values compared to the baselines. Higher recall values indicate that the algorithm is able to identify clean samples more reliably. This is useful, for example, to employ a label cleaning algorithm on the samples flagged as noisy (i.e., not selected) by BARE. CoT+ selects a fraction of samples where two networks disagree and, hence, after the first few epochs, it selects very few samples ($\sim$ 3000) in each epoch. Since these are samples in which the networks disagree, a good fraction of them may have noisy labels. This may be the reason for the poor precision and recall values of CoT+ as seen in these figures.

\begin{table}[h]
	\caption{Algorithm run times for training (in seconds)}
	\label{table:alg-time}
	%	\vskip 0.15in
	\begin{center}
		\begin{small}
			\begin{sc}
				\begin{tabular}{l|c|r}
					\toprule
					Algorithm & MNIST & CIFAR10 \\
					\midrule
					\textbf{BARE} & \textbf{310.64} & \textbf{930.78} \\
					CoT & 504.5 &  1687.9 \\
					CoT+ & 537.7 & 1790.57 \\
					MR & 807.4 & 8130.87 \\
					MN & 1138.4 & 8891.6 \\
					CL & 730.15 & 1254.3 \\
					\textbf{CCE} & \textbf{229.27} & \textbf{825.68} \\					
					\bottomrule
				\end{tabular}
			\end{sc}
		\end{small}
	\end{center}
	\vskip -0.1in
\end{table}

This can be seen from Figure \ref{fig:cifar10-cc-frac} as well which shows the fraction of samples chosen by the sample selection algorithms as epochs go by for $\eta = 0.4$ (class-conditional noise) on CIFAR-10 dataset. It can be noted that, as noise rate is to be supplied to CoT and CL, they select $1 - \eta = 0.6$ fraction of data with every epoch. Whereas, in case of CoT+, the samples where the networks disagree is small because of the training dynamics and as a result, after a few epochs, it consistently selects very few samples. Since the noise is class-conditional, even though $\eta = 0.4$, the actual amount of label flipping is $\sim20\%$. And this is why it's interesting to note that BARE leads to an approximate sample selection ratio of $80\%$. (We provide similar plots for different noise rates and datasets in the supplementary.)

\textbf{Efficiency of BARE.} Table \ref{table:alg-time} shows the typical run times for 200 epochs of training with all the algorithms. It can be seen from the table that the proposed algorithm takes roughly the same time as the usual training with CCE loss whereas all other baselines are significantly more expensive computationally. In case of MR and MN, the run times are around 8 times that of BARE for CIFAR-10. 

\textbf{Sensitivity to noise rates.} Some of the baselines schemes such as CoT, CoT+, and CL require knowledge of true noise rates beforehand. (In fact, in the simulations shown so far, we have used the actual noise rate for these baselines). This information is typically unavailable in practice. One can estimate the noise rates but there would be inevitable errors in estimation. Figure \ref{fig:mnist-cifar10-sent-samp-frac-2} shows the effect of mis-specification of noise rates for these 3 baselines schemes. As can be seen from these figures, while the algorithms can exhibit robust learning when the true noise rate is known, the performance deteriorates if the estimated noise rate is erroneous. BARE does not have this issue because it does not need any information on noise rate.

\textbf{Sensitivity to batch size.} To show the insensitivity to batch size, we show in Table~\ref{table:sensit-bs} results on MNIST \& CIFAR-10 for both types of label noise and three batch sizes: 64, 128 (used in paper), and 256.

\vspace*{-0.2cm}
\section{Conclusions}
\label{sec:conclusion}

\vspace*{-0.1cm}
We proposed an adaptive sample selection scheme, BARE, for robust learning under label noise. The algorithm relies on statistics of scores (posterior probabilities) of all samples in a minibatch to select samples from that minibatch. The current algorithms for sample selection in literature rely on heuristics such as cross-training multiple networks or meta-learning of sample weights which is often computationally expensive. They may also need knowledge of noise rates or some data with clean labels which may not be easily available. In contrast, BARE neither needs an extra data set with clean labels nor does it need any knowledge of the noise rates, nor does it need to learn multiple networks. Furthermore, it has no hyperparameters in the selection algorithm. Comparisons with baseline schemes on benchmark datasets show the effectiveness of the proposed algorithm both in terms of performance metrics and computational complexity. In addition, performance figures in terms of precision and recall show that BARE is very reliable in selecting clean samples. This, combined with the fact that there are no additional hyperparameters to tune, shows the advantage that BARE can offer for robust learning under label noise.

\begin{table}[h!]
	\caption{Test Accuracy (\%) of BARE on MNIST \& CIFAR-10 with batch sizes $ \in\{64,128,256\} $}
	\label{table:sensit-bs}
	\vspace*{-0.3cm}
	\begin{center}
		\begin{footnotesize}
			\begin{sc}
				\begin{tabular}{c|c|c|c}
					\toprule
					Dataset & Noise ($ \eta $) & Batch Size & Test Accuracy \\
					\midrule
					& & 64 & $ 95.31 \pm 0.16$ \\
					MNIST & 50\% (sym.) & 128 & $ 94.38\pm0.13 $ \\
					& & 256 & $94.44 \pm 0.48$ \\
					\midrule
					& & 64 & $ 93.31 \pm 0.63 $ \\
					MNIST & 45\% (cc) & 128 & $94.11\pm0.77$ \\
					& & 256 & $ 94.68 \pm 0.63 $ \\
					\midrule
					& & 64 & $ 76.77\pm0.38 $ \\
					CIFAR-10 & 30\% (sym.) & 128 & $ 75.85\pm0.41 $ \\
					& & 256 & $ 74.56\pm0.53 $ \\
					\midrule
					& & 64 & $71.87 \pm 0.28$ \\
					CIFAR-10 & 40\% (cc) & 128 & $ 70.63\pm0.46 $ \\
					& & 256 & $69.03 \pm 0.35$ \\
					\bottomrule
				\end{tabular}
			\end{sc}
		\end{footnotesize}
	\end{center}
	\vspace*{-0.6cm}
\end{table}

The mini-batch statistics used in BARE are class-specific. Hence, one may wonder whether such statistics would be reliable when the number of classes is large and hence is comparable to the mini-batch size. Our preliminary investigations show that the method delivers good performance even on a 101-class dataset with a minibatch size of 128 (see Table \ref{table:food-101} in Supplementary material). A possible approach for tackling large number of classes would be to make mini-batches in such a way that any given mini-batch contains examples of only a few of the classes (though for a full epoch there would be no class-imbalance). More investigations are needed to study this aspect of BARE.

%\begin{figure}[t]
%\begin{center}
%   %\includegraphics[width=0.8\linewidth]{egfigure.eps}
%\end{center}
%   \caption{Example of caption.  It is set in Roman so that mathematics
	%   (always set in Roman: $B \sin A = A \sin B$) may be included without an
	%   ugly clash.}
%\label{fig:long}
%\label{fig:onecol}
%\end{figure}
%\clearpage
{\small
\bibliographystyle{ieee_fullname}
\bibliography{references}

\twocolumn[\centering {\Large \textbf{Adaptive Sample Selection for Robust Learning under Label Noise - Supplementary Material\vspace{0.3cm}}}]

\subsection*{Network architectures \& Optimizers}

We use one MLP and one CNN architecture. For MNIST we train a 1-hidden layer fully-connected network with Adam (learning rate = $ 2\times10^{-4} $ and a learning rate scheduler: ReduceLROnPlateau). (This is same as the network used for Co-Teaching and Co-Teaching+ \cite{coteaching,coteaching+}). For CIFAR-10 we train a 4-layer CNN with Adam \cite{adam} (learning rate = $ 2\times10^{-3} $ and a learning rate scheduler: ReduceLROnPlateau). All networks are trained for 200 epochs. For MR, SGD optimizer with momentum 0.9 and learning rate of $1\times10^{-3}$ is used as the meta-optimizer. For MN, SGD optimizer with learning rate of $2\times10^{-3}$ is used as meta-optimizer. For CL, soft hinge loss is used as suggested in \cite{curr-loss} instead of cross-entropy loss. Rest of the algorithms use cross-entropy loss. All the simulations are run for 5 trials. A pre-trained ResNet-50 is used for training on Clothing-1M with SGD (learning rate of $1\times10^{-3}$ that is halved at epochs 6 and 11) with a weight decay of $1\times10^{-3}$ and momentum 0.9 for 14 epochs. All experiments use PyTorch \cite{pytorch}, NumPy \cite{numpy}, scikit-learn \cite{scikit-learn}, and NVIDIA Titan X Pascal GPU with CUDA 10.0. % For NEWS dataset, we train a 3-layer fully-connected network with pre-trained word embeddings from GloVe \cite{glove} as described in \cite{coteaching+} with Adam (learning rate = $ 1e-3 $) (and a learning rate scheduler: ??). 

These settings of optimizer, learning rate, and learning rate scheduler were found to work the best for our experimental and hardware setup. 

With our work for robust learning under label noise, our objective has been to show the degree of robustness offered by BARE for any given architecture. Keeping this in mind along with maintaining some commonality with the architectures used for baseline schemes (i.e., their corresponding papers), we have shown simulation results for the aforementioned MLPs and 4-layer CNNs. However, it should be noted that BARE continues to demonstrate its effectiveness on bigger architectures such as ResNet-50 as seen in Clothing-1M performance results (Table \ref{table:clothing-1m}).

\subsection*{Test accuracies on MNIST \& CIFAR-10}
We tabulate the test accuracies of all the algorithms on MNIST and CIFAR-10 in Tables \ref{table:mnist-sym-05-acc} -- \ref{table:cifar10-cc-04-acc}. The best two results are in bold. These are accuracies achieved at the end of training. For CoT \cite{coteaching} and CoT+ \cite{coteaching+}, we show accuracies only of that network which performs the best out of the two that are trained. In the main paper we showed the plots of accuracies.

It may be noted that in a couple of cases the standard deviation in the accuracy for MN is high. As we mentioned in the main paper, we noticed that MN is very sensitive to the tuning of hyper parameters. While we tried our best to tune all the hyper parameters, maybe the final ones we found for these two cases are still not the best and that is why the standard deviation is high. 

%\begin{savenotes}
	\begin{table}[ht]
		\caption{Test accuracies on Clothing-1M dataset}
		\label{table:clothing-1m}
		%	\vskip 0.15in
		\begin{center}
%			\begin{small}
%				\begin{sc}
					\begin{tabular}{l|r}
						\toprule
						Algorithm & Test Accuracy (\%) \\
						\midrule
						CCE & 68.94 \\
						D2L \cite{d2l} & 69.47 \\
						GCE \cite{gce} & 69.75 \\					
						Forward \cite{loss-correction} & 69.84 \\
						CoT \cite{coteaching} (as reported in \cite{seal}) & 70.15 \\			
						JoCoR \cite{jocor} & 70.30 \\
						SEAL \cite{seal} & 70.63 \\
						DY \cite{arazo2019unsupervised} & 71.00 \\
						SCE \cite{sym-ce} & 71.02 \\
						LRT \cite{error-bound-icml20} & 71.74 \\
						PTD-R-V \cite{inst-dep-sugiyama} & 71.67 \\
						Joint Opt. \cite{joint-opt} & 72.23 \\
						\textbf{BARE (Ours)} & \textbf{72.28} \\
						C2D \cite{contrast-self-2022} {\footnotesize (ELR+ \cite{elr-plus} with SimCLR \cite{SimCLR})} &  74.58 \\
						DivideMix \cite{dividemix} & 74.76 \\
						\bottomrule
					\end{tabular}
%				\end{sc}
%			\end{small}
		\end{center}
		\vskip -0.1in
	\end{table}
%\end{savenotes}

\begin{table}[h]
	\caption{Test Accuracy (\%) for MNIST - $\eta = 0.5$ (symmetric)}
	\label{table:mnist-sym-05-acc}
	\vskip 0.15in
	\begin{center}
		%		\begin{small}
			\begin{sc}
				\begin{tabular}{l|cr}
					\toprule
					Algorithm & Test Accuracy \\
					\midrule
					CoT \cite{coteaching} & $90.80\pm0.18$ \\
					CoT+ \cite{coteaching+}	& $\boldsymbol{93.17\pm0.3}$ \\
					MR \cite{meta-ren}  & $90.39\pm0.07$\\
					MN \cite{meta-net} & $74.94\pm9.56$\\
					CL \cite{curr-loss} & $92.00\pm0.26$ \\
					CCE & $74.30\pm0.55$\\
					\midrule
					\textbf{BARE (Ours)} & $\boldsymbol{94.38\pm0.13}$ \\
					\bottomrule
				\end{tabular}
			\end{sc}
			%		\end{small}
	\end{center}
	\vskip -0.1in
\end{table}
\begin{table}[h]
	\caption{Test Accuracy (\%) for MNIST - $\eta = 0.7$ (symmetric)}
	\label{table:mnist-sym-07-acc}
	\vskip 0.15in
	\begin{center}
		%		\begin{small}
			\begin{sc}
				\begin{tabular}{l|cr}
					\toprule
					Algorithm & Test Accuracy \\
					\midrule
					CoT \cite{coteaching} & $87.17\pm0.45$ \\
					CoT+ \cite{coteaching+}	& $87.26\pm0.67$ \\
					MR \cite{meta-ren}  & $85.10\pm0.28$ \\
					MN \cite{meta-net} & $65.52\pm21.35$ \\
					CL \cite{curr-loss} & $\boldsymbol{88.28\pm0.45}$ \\
					CCE & $61.19\pm1.29$\\
					\midrule
					\textbf{BARE (Ours)} & $\boldsymbol{91.61\pm0.60}$ \\
					\bottomrule
				\end{tabular}
			\end{sc}
			%		\end{small}
	\end{center}
	\vskip -0.1in
\end{table}

\begin{table*}[h]
	\caption{Test Accuracy (\%) for MNIST - $\eta = 0.45$ (class-conditional)}
	\label{table:mnist-cc-045-acc}
	\vskip 0.15in
	\begin{center}
		%		\begin{small}
			\begin{sc}
				\begin{tabular}{l|cr}
					\toprule
					Algorithm & Test Accuracy \\
					\midrule
					CoT \cite{coteaching} & $\boldsymbol{95.20\pm0.22}$ \\
					CoT+ \cite{coteaching+}	& $91.10\pm1.51$ \\
					MR \cite{meta-ren}  & $\boldsymbol{95.40\pm0.31}$ \\
					MN \cite{meta-net} & $75.03\pm0.59$ \\
					CL \cite{curr-loss} & $81.52\pm3.27$ \\
					CCE & $74.96\pm0.21$\\
					\midrule
					\textbf{BARE (Ours)} & $94.11\pm0.77$ \\
					\bottomrule
				\end{tabular}
			\end{sc}
			%		\end{small}
	\end{center}
	\vskip -0.1in
\end{table*}
\begin{table*}[h]
	\caption{Test Accuracy (\%) for CIFAR-10 - $\eta = 0.3$ (symmetric)}
	\label{table:cifar10-sym-03-acc}
	\vskip 0.15in
	\begin{center}
		%		\begin{small}
			\begin{sc}
				\begin{tabular}{l|cr}
					\toprule
					Algorithm & Test Accuracy \\
					\midrule
					CoT \cite{coteaching} & $\boldsymbol{71.72\pm0.30}$ \\
					CoT+ \cite{coteaching+}	& $60.14\pm0.35$ \\
					MR \cite{meta-ren}  & $62.96\pm0.70$ \\
					MN \cite{meta-net} & $51.65\pm1.49$ \\
					CL \cite{curr-loss} & $66.124\pm0.45$ \\
					CCE & $54.83\pm0.28$\\					
					\midrule
					\textbf{BARE (Ours)} & $\boldsymbol{75.85\pm0.41}$ \\
					\bottomrule
				\end{tabular}
			\end{sc}
			%		\end{small}
	\end{center}
	\vskip -0.1in
\end{table*}
\begin{table*}[h]
	\caption{Test Accuracy (\%) for CIFAR-10 - $\eta = 0.7$ (symmetric)}
	\label{table:cifar10-sym-07-acc}
	\vskip 0.15in
	\begin{center}
		%		\begin{small}
			\begin{sc}
				\begin{tabular}{l|cr}
					\toprule
					Algorithm & Test Accuracy \\
					\midrule
					CoT \cite{coteaching} & $\boldsymbol{58.95\pm1.31}$ \\
					CoT+ \cite{coteaching+}	& $37.69\pm0.70$ \\
					MR \cite{meta-ren}  & $45.14\pm1.04$ \\
					MN \cite{meta-net} & $23.23\pm0.65$ \\
					CL \cite{curr-loss} & $44.82\pm2.42$ \\
					CCE & $23.46\pm0.37$\\
					\midrule
					\textbf{BARE (Ours)} & $\boldsymbol{59.53\pm1.12}$ \\
					\bottomrule
				\end{tabular}
			\end{sc}
			%		\end{small}
	\end{center}
	\vskip -0.1in
\end{table*}
\begin{table*}[h]
	\caption{Test Accuracy (\%) for CIFAR-10 - $\eta = 0.4$ (class-conditional)}
	\label{table:cifar10-cc-04-acc}
	\vskip 0.15in
	\begin{center}
		%		\begin{small}
			\begin{sc}
				\begin{tabular}{l|cr}
					\toprule
					Algorithm & Test Accuracy \\
					\midrule
					CoT \cite{coteaching} & $65.26\pm0.78$ \\
					CoT+ \cite{coteaching+}	& $63.05\pm0.39$ \\
					MR \cite{meta-ren}  & $\boldsymbol{70.27\pm0.77}$ \\
					MN \cite{meta-net} & $63.84\pm0.41$ \\
					CL \cite{curr-loss} & $64.48\pm2.02$ \\
					CCE & $64.06\pm0.32$\\
					\midrule
					\textbf{BARE (Ours)} & $\boldsymbol{70.63\pm0.46}$ \\
					\bottomrule
				\end{tabular}
			\end{sc}
			%		\end{small}
	\end{center}
	\vskip -0.1in
\end{table*}

\subsection*{Performance on Clothing-1M dataset}

Table \ref{table:clothing-1m} shows how the proposed algorithm fares against several baselines in terms of test accuracy on Clothing-1M dataset. For the baselines, we report the accuracy values as reported in the corresponding papers.

\clearpage

\subsection*{Plots for sample fraction v/s epochs}
We show some more plots for fraction of samples chosen by the baselines along with the proposed scheme, BARE, as epochs go by for some more dataset and label noise combinations in Figures \ref{fig:mnist-samp-frac-sym-05} \& \ref{fig:cifar10-samp-frac-sym-07}. As noted during the discussion in the paper, and as is evident from these figures, BARE is able to identify the clean samples effectively even without the knowledge of noise rates.

\begin{figure}[h]
	\centering
	\includegraphics[scale=0.04]{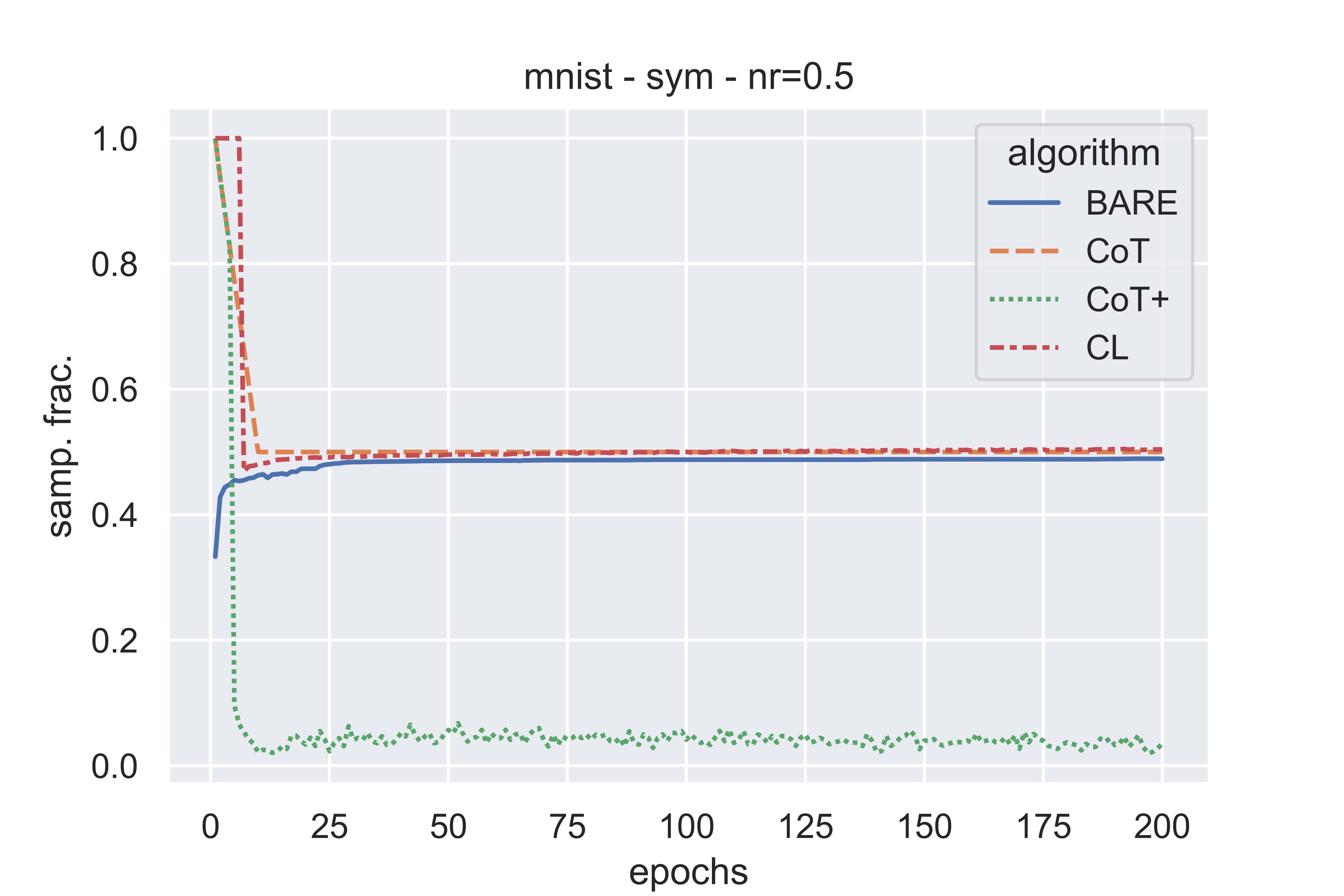}
	\caption{Sample fraction values for $\eta = 0.5$ (symmetric noise) on MNIST}
	\label{fig:mnist-samp-frac-sym-05}	
\end{figure}

\begin{figure}[h]
	\centering
	\includegraphics[scale=0.04]{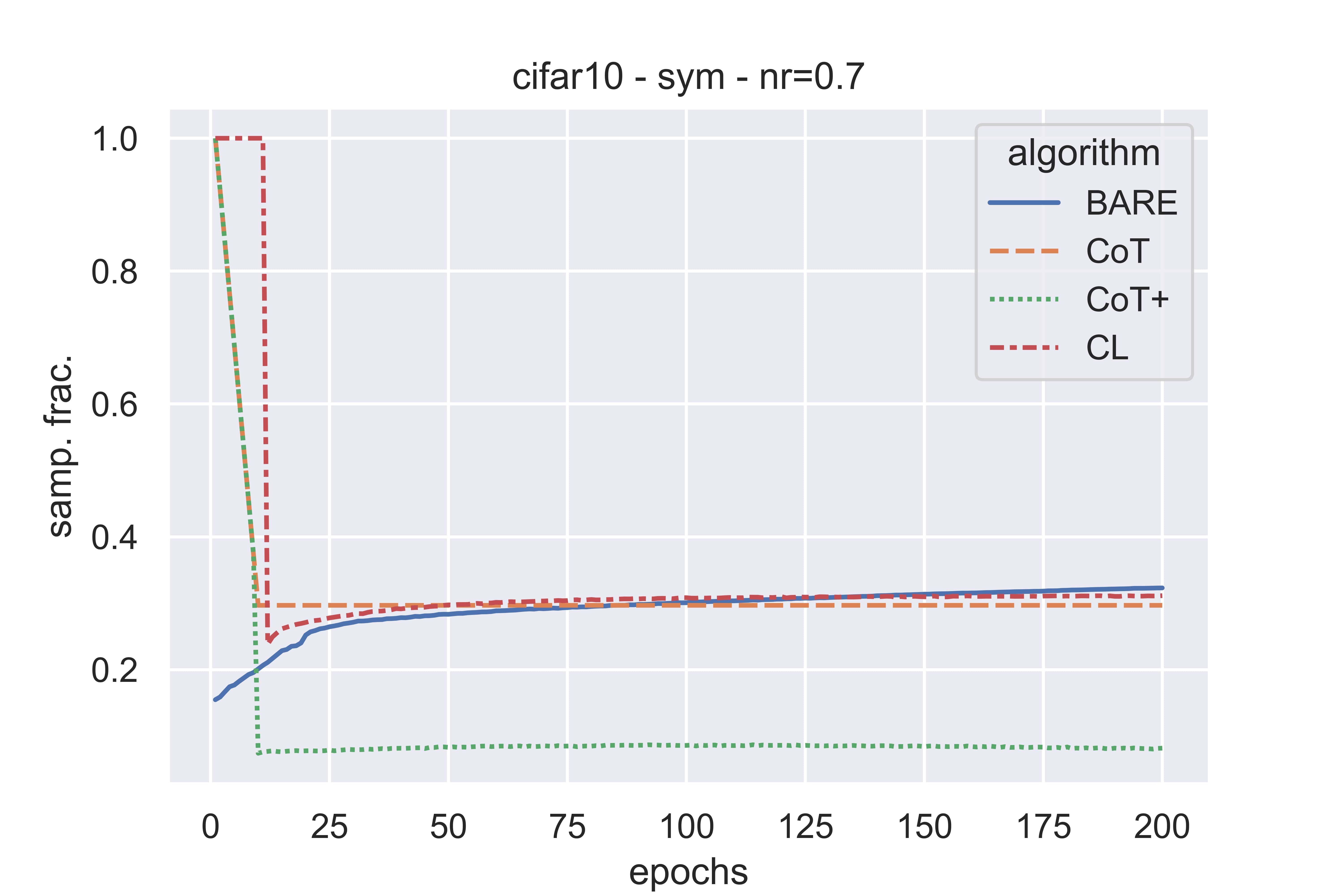}
	\caption{Sample fraction values for $\eta = 0.7$ (symmetric noise) on CIFAR-10}
	\label{fig:cifar10-samp-frac-sym-07}	
\end{figure}

\vspace{-0.25cm}

\subsection*{Results on Arbitrary Noise Transition Matrix}
In the main paper, we showed results for special class-conditional noise cases taken from literature. Here we provide results for an arbitrary, digonally-dominant noise transition matrix in Tables \ref{table:mnist-arbit-mat-perf}--\ref{table:cifar10-arbit-mat-perf-l10}. $\eta=0.45$ and $\eta=0.4$ are supplied as the estimated noise rates to CoT, CoT+, and CL baselines for MNIST and CIFAR-10 respectively. The best two results are in bold. It can be seen that the proposed algorithm continues to perform well. The noise transition matrices are arbitrary but for the sake of completeness we show them at the end of this supplementary material.

\begin{table}[h]
	\caption{Test Accuracy (\%) for MNIST - $\eta_{est} = 0.45$ (arbitrary noise matrix)}
	\label{table:mnist-arbit-mat-perf}
	\vskip 0.15in
	\begin{center}
		\begin{small}
			\begin{sc}
				\begin{tabular}{l|cr}
					\toprule
					Algorithm & Test Accuracy\\
					\midrule
					CoT \cite{coteaching} & $\boldsymbol{95.3}$\\
					CoT+ \cite{coteaching+}	& 93.07\\
					CL \cite{curr-loss} & 88.41\\
					\midrule
					\textbf{BARE (Ours)} & $\boldsymbol{95.02}$\\
					\bottomrule
				\end{tabular}
			\end{sc}
		\end{small}
	\end{center}
	\vskip -0.1in
\end{table}

\begin{table}[h]
	\caption{Avg. Test Accuracy (last 10 epochs) (\%) for MNIST - $\eta_{est} = 0.45$ (arbitrary noise matrix)}
	\label{table:mnist-arbit-mat-perf-l10}
	\vskip 0.15in
	\begin{center}
		\begin{small}
			\begin{sc}
				\begin{tabular}{l|cr}
					\toprule
					Algorithm & Avg. Test Accuracy (last 10 epochs)\\
					\midrule
					CoT \cite{coteaching} & $\boldsymbol{95.22}$ \\
					CoT+ \cite{coteaching+} & 93.08\\
					CL \cite{curr-loss} & 88.56 \\
					\midrule
					\textbf{BARE (Ours)} & $\boldsymbol{95.03}$ \\
					\bottomrule
				\end{tabular}
			\end{sc}
		\end{small}
	\end{center}
	\vskip -0.1in
\end{table}

\begin{table}[h]
	\caption{Test Accuracy (\%) for CIFAR10 - $\eta_{est} = 0.4$ (arbitrary noise matrix)}
	\label{table:cifar10-arbit-mat-perf}
	\vskip 0.15in
	\begin{center}
		\begin{small}
			\begin{sc}
				\begin{tabular}{l|cr}
					\toprule
					Algorithm & Test Accuracy\\
					\midrule
					CoT \cite{coteaching} & 71.92\\
					CoT+ \cite{coteaching+}	& 68.56\\
					CL \cite{curr-loss} & $\boldsymbol{72.12}$\\
					\midrule
					\textbf{BARE (Ours)} & $\boldsymbol{76.22}$\\
					\bottomrule
				\end{tabular}
			\end{sc}
		\end{small}
	\end{center}
	\vskip -0.1in
\end{table}

\begin{table}[h]
	\caption{Avg. Test Accuracy (last 10 epochs) (\%) for CIFAR10 - $\eta_{est} = 0.4$ (arbitrary noise matrix)}
	\label{table:cifar10-arbit-mat-perf-l10}
	\vskip 0.15in
	\begin{center}
				\begin{small}
			\begin{sc}
				\begin{tabular}{l|cr}
					\toprule
					Algorithm & Avg. Test Accuracy (last 10 epochs)\\
					\midrule
					CoT \cite{coteaching} & 71.86 \\
					CoT+ \cite{coteaching+} & 68.99\\
					CL \cite{curr-loss} & $\boldsymbol{72.27}$ \\
					\midrule
					\textbf{BARE (Ours)} & $\boldsymbol{75.96}$ \\
					\bottomrule
				\end{tabular}
			\end{sc}
					\end{small}
	\end{center}
	\vskip -0.1in
\end{table}

\newpage

\begin{table*}[h]
	\caption{Network Architectures used for training on MNIST and CIFAR-10 datasets}
	\label{table:arch}
	\vskip 0.15in
	\begin{center}
		\begin{small}
			\begin{sc}
				\begin{tabular}{c|c}
					\toprule
					MNIST & CIFAR-10 \\
					\midrule
					\multirow{11}{*}{dense 28$\times$28 $\rightarrow$ 256} & 3$\times$3 conv., 64 ReLU, stride 1, padding 1\\
					\cline{2-2} 
					& Batch Normalization\\
					\cline{2-2} 
					& 2$\times$2 Max Pooling, stride 2\\
					\cline{2-2} 
					& 3$\times$3 conv., 128 ReLU, stride 1, padding 1\\
					\cline{2-2} 
					& Batch Normalization\\
					\cline{2-2} 
					& 2$\times$2 Max Pooling, stride 2\\
					\cline{2-2} 
					& 3$\times$3 conv., 196 ReLU, stride 1, padding 1\\
					\cline{2-2} 
					& Batch Normalization\\
					\cline{2-2} 
					& 3$\times$3 conv., 16 ReLU, stride 1, padding 1\\
					\cline{2-2} 
					& Batch Normalization\\
					\cline{2-2} 
					& 2$\times$2 Max Pooling, stride 2\\
					\hline 
					dense 256 $\rightarrow$ 10 & dense 256 $\rightarrow$10\\
					\hline
				\end{tabular}
			\end{sc}
		\end{small}
	\end{center}
	\vskip -0.1in
\end{table*}

\newpage

\begin{figure*}
	\centering
	$	\begin{bmatrix}
		1 & 0 & 0 & 0 & 0 & 0 & 0 & 0 & 0 & 0\\
		0 & 1 & 0 & 0 & 0 & 0 & 0 & 0 & 0 & 0\\
		0 & 0 & 0.6 & 0 & 0 & 0 & 0 & 0.3 & 0 & 0.1\\
		0 & 0 & 0 & 0.5 & 0 & 0.1 & 0 & 0 & 0.4 & 0\\
		0 & 0 & 0 & 0 & 1 & 0 & 0 & 0 & 0 & 0\\
		0 & 0 & 0 & 0 & 0.15 & 0.55 & 0.3 & 0 & 0 & 0\\
		0 & 0 & 0 & 0 & 0 & 0.35 & 0.55 & 0.10 & 0 & 0\\
		0 & 0.25 & 0 & 0 & 0 & 0 & 0 & 0.5 & 0 & 0.25\\
		0 & 0 & 0 & 0 & 0 & 0 & 0 & 0 & 1 & 0\\
		0 & 0 & 0 & 0 & 0 & 0 & 0 & 0 & 0 & 1\\
	\end{bmatrix} $
	\caption*{Arbitrary Noise Transition Matrix for MNIST}
\end{figure*}

\begin{figure*}
	\centering
	$ \begin{bmatrix}
		1 & 0 & 0 & 0 & 0 & 0 & 0 & 0 & 0 & 0\\
		0 & 1 & 0 & 0 & 0 & 0 & 0 & 0 & 0 & 0\\
		0.2 & 0 & 0.7 & 0 & 0 & 0 & 0.1 & 0 & 0 & 0\\
		0.1 & 0 & 0 & 0.6 & 0 & 0.1 & 0 & 0 & 0.2 & 0\\
		0 & 0.1 & 0.1 & 0 & 0.7 & 0 & 0 & 0.1 & 0 & 0\\
		0 & 0 & 0 & 0.1 & 0 & 0.6 & 0 & 0 & 0 & 0.3\\
		0 & 0 & 0 & 0 & 0 & 0 & 1 & 0 & 0 & 0\\
		0 & 0 & 0 & 0 & 0 & 0 & 0 & 1 & 0 & 0\\
		0 & 0 & 0 & 0 & 0 & 0 & 0 & 0 & 1 & 0\\
		0 & 0.1 & 0 & 0 & 0 & 0 & 0 & 0.1 & 0 & 0.8\\
	\end{bmatrix} $
	\caption*{Arbitrary Noise Transition Matrix for CIFAR-10}
\end{figure*}

\clearpage

\section*{Performance under different noise rates}
The proposed algorithm shows fairly stable robustness across noise rates. The paper reports results on three noise rates in Figures 1 \& 2. Table 2 is about the effect of minibatch size and we included a different noise rate here. We report some more results on different noise rates. Table \ref{table:mnist-more-nr} shows performance of BARE on MNIST under different noise rates (averaged over 5 runs of 200 epochs each) and Table \ref{table:sensit-bs-more} shows some more empirical results on sensitivity of BARE to batch-sizes.

\begin{table}[h]
	\caption{Test Accuracy (\%) of BARE on MNIST \& CIFAR-10}
	\label{table:mnist-more-nr}
	\vskip 0.15in
	\begin{center}
				\begin{small}
			\begin{sc}
				\begin{tabular}{l|c|r}
					\toprule
					Dataset & Noise ($\eta$) & Test Accuracy (in \%) \\
					\midrule
					& 10\% (sym.) & $96.7\pm0.31$ \\
					& 20\% (sym.)	& $96.39\pm0.23$ \\
					& 30\% (sym.)  & $ 95.92\pm0.17 $ \\
					& 40\% (sym.) & $ 95.22\pm 0.23 $\\
					MNIST & 50\% (sym.) & $94.38\pm0.13$\\
					& 60\% (sym.) & $93.44\pm0.27$\\
					& 70\% (sym.) & $91.61\pm0.59$\\
					& 20\% (cc) & $ 96.45 \pm 0.33 $\\
					& 45\% (cc) & $ 95.05 \pm $ 0.23 \\
					\midrule
					& 10\% (sym.) & $ 78.76\pm 0.42 $ \\
					& 20\% (sym.) & $ 77.13\pm 0.46 $ \\
					& 30\% (sym.) & $ 75.85\pm 0.41 $ \\
					CIFAR-10 & 40\% (sym.) & $ 73.86\pm 0.49 $ \\
					& 60\% (sym.) & $ 66.87\pm 0.82 $ \\
					& 70\% (sym.) & $ 59.87\pm 1.12 $ \\
					& 20\% (cc) & $ 76.78\pm 0.38 $ \\
					& 40\% (cc) & $ 70.63\pm 0.46 $ \\
					\bottomrule
				\end{tabular}
			\end{sc}
					\end{small}
	\end{center}
	\vskip -0.1in
\end{table}

\begin{table}
	\caption{Test Accuracy (\%) of BARE on MNIST \& CIFAR-10 with batch sizes $ \in\{64,128,256\} $}
	\label{table:sensit-bs-more}
	\vskip 0.15in
	\begin{center}
		\begin{footnotesize}
			\begin{sc}
				\begin{tabular}{c|c|c|c}
					\toprule
					Dataset & Noise ($ \eta $) & Batch Size & Test Accuracy \\
					\midrule
					& & 64 & $ 96.26 \pm 0.10$ \\
					MNIST & 30\% (sym.) & 128 & $ 96.03\pm0.20 $ \\
					& & 256 & $95.62 \pm 0.13$ \\
					\midrule
					& & 64 & $ 92.42 \pm 0.40 $ \\
					MNIST & 70\% (sym.) & 128 & $91.61\pm0.60$ \\
					& & 256 & $ 91.4 \pm 0.38 $ \\
					\midrule
					\midrule
					& & 64 & $ 59.97 \pm 0.66 $ \\
					CIFAR-10 & 70\% (sym.) & 128 & $ 59.53 \pm 1.12 $ \\
					& & 256 & $ 56.72 \pm 0.72 $ \\
					\bottomrule
				\end{tabular}
			\end{sc}
		\end{footnotesize}
	\end{center}
	\vskip -0.2in
\end{table}

\section*{Rationale for choosing baselines in this work}
\vspace*{-0.07cm}
There are now a very large number of algorithms motivated by different ideas for learning under label noise. Hence, we compared against only those methods that use sample selection as the main strategy and for these comparisons we used the same network for all methods and used the same noisy training set (with MNIST, CIFAR). For Clothing1M data, the results shown in supplementary material for different algorithms are all as reported in the respective papers and different results are with different network architectures. Here we reported results available in literature for different algorithms, including algorithms that do not rely on sample selection.

\section*{Regarding early-stopping and long training}
From Figures 1 \& 2, it is easily seen that irrespective of which epoch you would stop the other algorithms, the final accuracy achieved by BARE is better. For BARE, we can run the algorithm without any such necessity of early stopping. Further, we ran other algorithms that are close to BARE for upto 500 epochs and noticed that the test accuracies either saturated at the level shown in the Figures 1 \& 2 or decreased. See Table \ref{table:500-ep}. Apart from this, to be able to decide some epochs after which to stop, one needs clean validation data; the proposed algorithm does not need any such validation data.

%\vspace*{-0.5cm}

\begin{table}
	\caption{Test accuracies on MNIST under label noise for 500 epoch runs}
	\label{table:500-ep}
	
	%	\vskip 0.15in
	\begin{center}
		\begin{small}
			\begin{sc}
				\begin{tabular}{c|c|c|r}
					\toprule
					Dataset & Noise ($ \eta $) & Algorithm & Test Accuracy \\
					\midrule
					& & CoT \cite{coteaching} & $84.65 \pm 0.22$ \\
					MNIST & 70\% (sym.) & CoT+ \cite{coteaching+} & $ 79.52 \pm 0.91 $ \\
					& & MR \cite{meta-ren} & $ 80.36 \pm 0.56 $ \\
					& & BARE (Ours) & $ 91.52 \pm 0.61 $ \\
					\midrule
					& & CoT \cite{coteaching} & $95.33 \pm 0.14$ \\
					MNIST & 45\% (cc) & CoT+ \cite{coteaching+} & $ 84.69 \pm 1.81 $ \\
					& & MR \cite{meta-ren} & $ 93.92 \pm 0.20 $ \\
					& & BARE (Ours) & $ 94.65 \pm 0.57 $ \\
					\midrule
					\midrule
					& & CoT \cite{coteaching} & $58.08 \pm 1.61$ \\
					CIFAR-10 & 70\% (sym.) & CoT+ \cite{coteaching+} & $ 28.18 \pm 0.57 $ \\
					& & MR \cite{meta-ren} & $ 40.52 \pm 1.62 $ \\
					& & BARE (Ours) & $ 58.03 \pm 1.01 $ \\
					\midrule
					& & CoT \cite{coteaching} & $65.19 \pm 0.61$ \\
					CIFAR-10 & 40\% (cc) & CoT+ \cite{coteaching+} & $ 62.61 \pm 0.48 $ \\
					& & MR \cite{meta-ren} & $ 68.84 \pm 0.58 $ \\
					& & BARE (Ours) & $ 69.88 \pm 0.44 $ \\
					\bottomrule
				\end{tabular}
			\end{sc}
		\end{small}
	\end{center}
	\vskip -0.1in
	
\end{table}

%\clearpage

\section*{Performance with large number of classes}

Food-101N \cite{cleannet} is a dataset for food classification that consists of 101 classes and 310k training images collected from the web. Estimate label noise rate is $ \sim 20\% $ and it is feature-dependent noise. The Food-101 \cite{food101} testing set is used for testing, which contains 25k cleanly-labelled images. Following the baselines, we train a ResNet-50 network (pre-trained on ImageNet) for 30 epochs with SGD optimizer. The batch size taken is 128 and the initial learning rate is 0.01, which is divided by 10 every 10 epochs. For data-augmentation, we follow same procedure as baselines -- random horizontal flip, and resizing the image with a short edge of 256 and then randomly cropping a $ 224 \times 224 $ patch from the resized image. Table \ref{table:food-101} shows how BARE compares with the baselines. With randomly formed minibatches of size 128 (and without any regard for sample sizes of different classes in a minibatch), we get an accuracy of 84.12\% (
denoted as \textit{without batch-balance}) which is better than CleanNet (83.95\%) but not PLC (85.28\%).

\begin{table}[h]
	\caption{Test Accuracy (\%) on Food-101N dataset}
	\label{table:food-101}
	%	\vskip -0.2in
	\begin{center}
		%		\begin{small}
			\begin{sc}
				\begin{tabular}{l|cr}
					\toprule
					Algorithm & Test Accuracy \\
					\midrule
					CCE (standard) & 81.67\% \\
					CleanNet \cite{cleannet}	& 83.95\% \\
					\textbf{PLC \cite{zhang2020learning}}  & $\boldsymbol{85.28}$ \\
					\midrule
					\textbf{BARE (without batch-balance)} & $\boldsymbol{84.12}$ \\
					\bottomrule
				\end{tabular}
			\end{sc}
			%		\end{small}
	\end{center}
	\vskip -0.2in
\end{table}

\section*{Rationale for choosing threshold (in BARE)}
The main idea in the algorithm is to use minibatch statistics to derive the threshold on probabilities for sample selection. As is to be expected, it is effective when we choose samples from the `top quantiles'. We now show results obtained when the threshold is taken as ``$ \text{mean}+ \kappa * \text{std. dev.} $'' for $ \kappa \in \{-1, 0, 0.5, 0.75, 1, 1.25, 1.5\} $. The method works well for all values of $\kappa$ except $\kappa \in \{-1, 0\}$. These results are shown here in Table \ref{table:thresh-abl-bare}. We also noticed that the numerical value of the threshold varies across minibatches and is different for different clases thus justifying the motivation we started with. Figures \ref{fig:mnist-cls-thresh-sym} -- \ref{fig:cifar-cls-thresh-cc} show the evolution of these threshold values (i.e. RHS of Equation \ref{eq:thresh} with $ \kappa = 1 $) for some of the randomly picked classes in MNIST and CIFAR-10 dataset at $ 0^{th} $, $ 100^{th} $ and $ 200^{th} $ epoch under different noise rates and noise types.
 
\begin{table*}[h!]
	\caption{Test accuracies of BARE under label noise for $ \kappa \in \{-1, 0, 0.5, 0.75, 1, 1.25, 1.5\} $}
	\label{table:thresh-abl-bare}
	
	%		\vskip -0.15in
	\begin{center}
		\begin{large}
			\begin{sc}
				\begin{tabular}{l|c|c|r}
					\toprule
					Dataset & Noise ($ \eta $) & $\kappa$ & Test Accuracy \\
					\midrule
					& & -1 & $58.25 \pm 0.73$ \\
					& & 0 & $ 92.1 \pm 0.76 $ \\
					& & 0.5 & $ 91.32 \pm 0.82 $ \\
					MNIST & 70\% (sym.) & 0.75 & $ 91.37 \pm 0.37 $ \\
					& & 1 & $ 91.61 \pm 0.59 $ \\
					& & 1.25 & $ 91.31 \pm 0.32 $ \\
					& & 1.5 & $ 91.2 \pm 0.53 $ \\
					\midrule
					& & -1 & $74.95 \pm 0.39$ \\
					& & 0 & $ 93.8 \pm 1.54 $ \\
					& & 0.5 & $ 94.26 \pm 1.06 $ \\
					MNIST & 45\% (cc) & 0.75 & $ 93.06 \pm 1.65 $ \\
					& & 1 & $ 94.11 \pm 0.77 $ \\
					& & 1.25 & $ 93.2 \pm 2.84 $ \\
					& & 1.5 & $ 92.72 \pm 3.71 $ \\
					\midrule
					\midrule
					& & -1 & $23.3 \pm 0.33$ \\
					& & 0 & $ 42.49 \pm 3.44 $ \\
					& & 0.5 & $ 58.17 \pm 0.50 $ \\
					CIFAR-10 & 70\% (sym.) & 0.75 & $ 59.04 \pm 0.67 $ \\
					& & 1 & $ 59.93 \pm 1.12 $ \\
					& & 1.25 & $ 57.14 \pm 1.78 $ \\
					& & 1.5 & $ 56.66 \pm 0.66 $ \\
					\midrule
					& & -1 & $63.77 \pm 0.11$ \\
					& & 0 & $ 64.70 \pm 0.31 $ \\
					& & 0.5 & $ 67.92 \pm 0.29 $ \\
					CIFAR-10 & 40\% (cc) & 0.75 & $ 69.87 \pm 0.34 $ \\
					& & 1 & $ 70.63 \pm 0.46 $ \\
					& & 1.25 & $ 71.06 \pm 0.56 $ \\
					& & 1.5 & $ 70.81 \pm 1.22 $ \\
					\bottomrule
				\end{tabular}
			\end{sc}
		\end{large}
	\end{center}
	\vskip -0.1in
\end{table*}

%\vspace*{-0.5cm}

\clearpage

\begin{figure*}
	\hspace*{0.8cm}\subfloat[]{\label{fig:mnist-cls-0-sym}\includegraphics[scale=0.04]{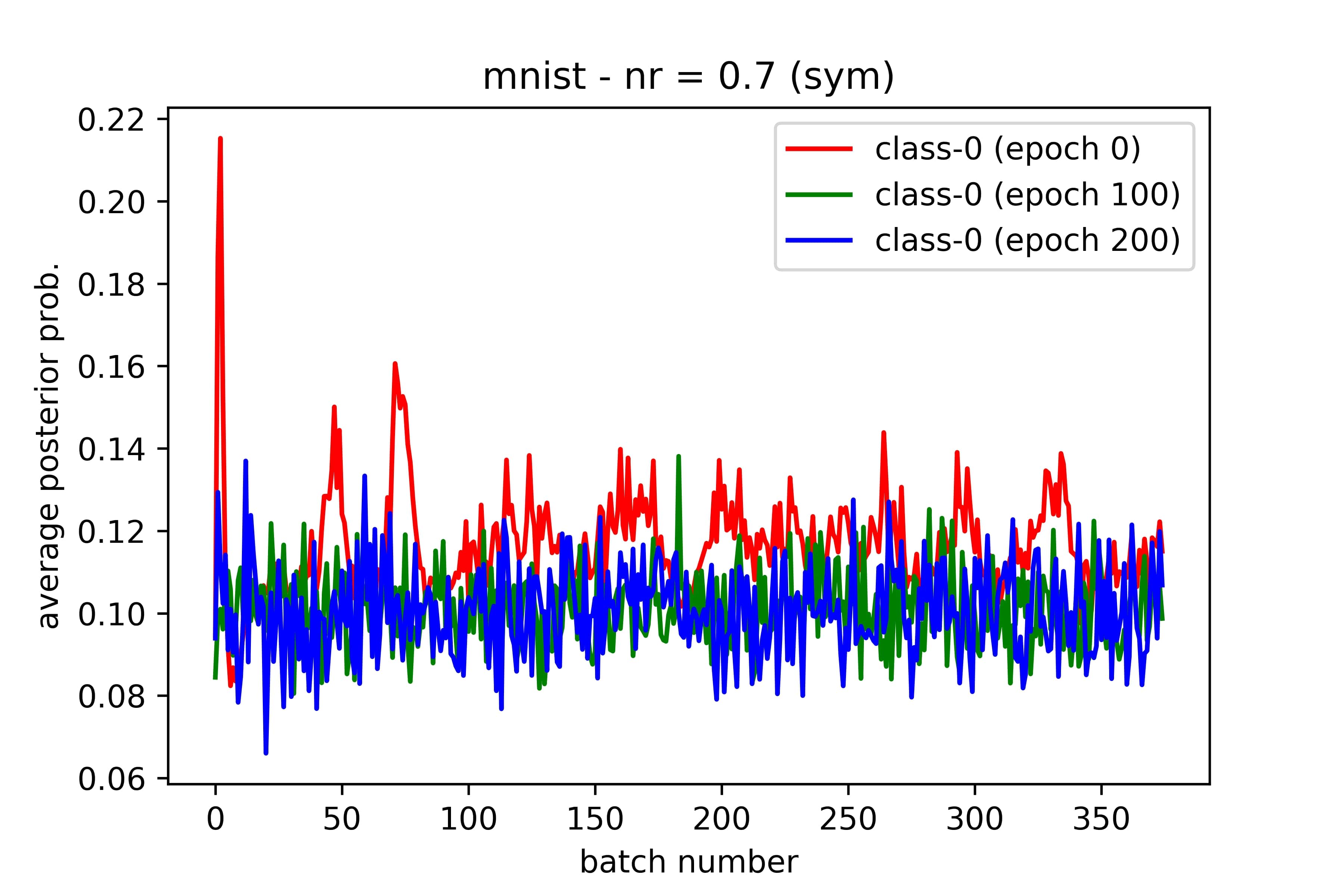}} \subfloat[]{\label{fig:mnist-cls-3-sym}\includegraphics[scale=0.04]{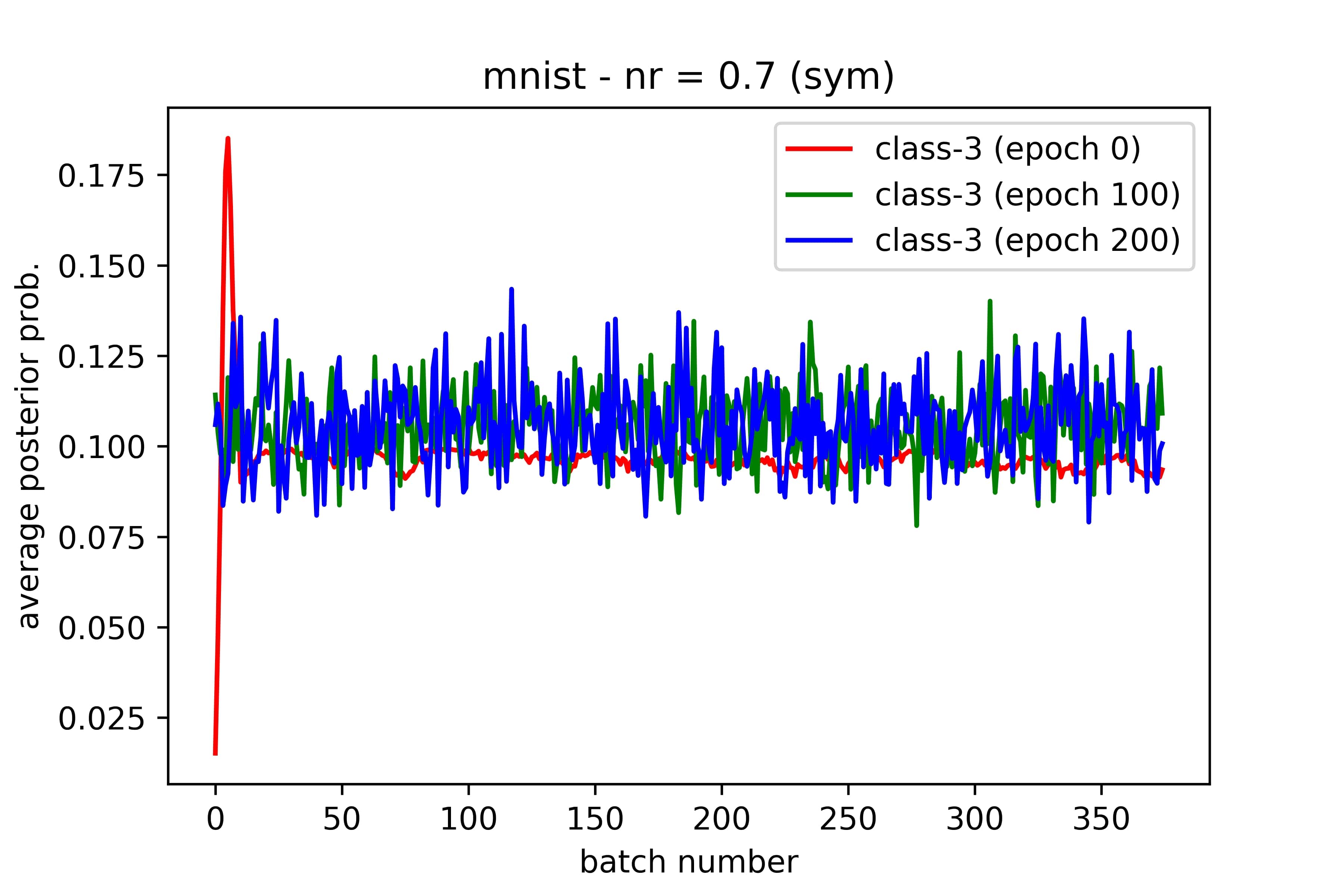}}
	\subfloat[]{\label{fig:mnist-cls-4-sym}\includegraphics[scale=0.04]{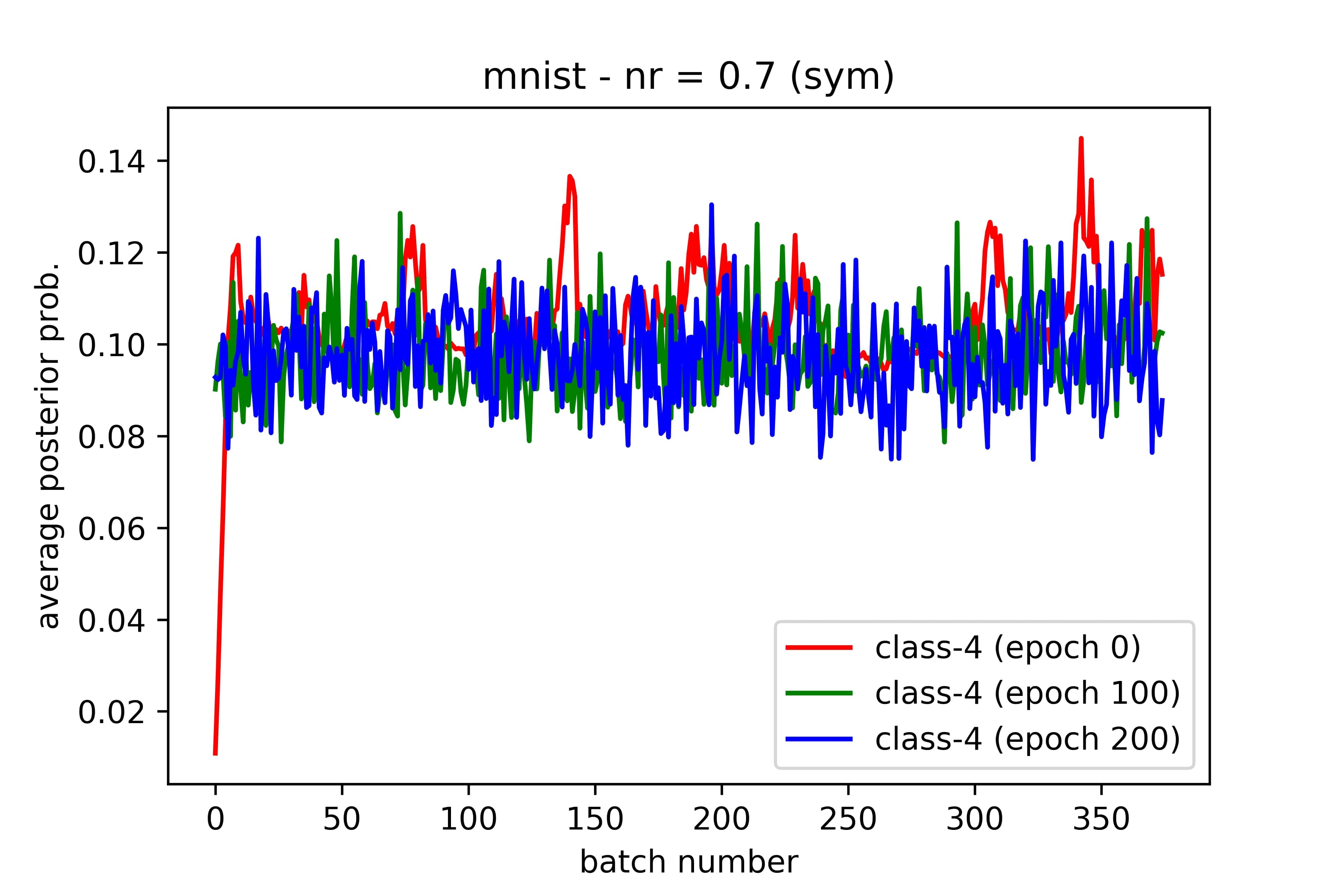}}
	
	\caption{Average class-wise posterior probability across mini-batches; (\subref{fig:mnist-cls-0-sym}): class 0; (\subref{fig:mnist-cls-3-sym}): class 3; (\subref{fig:mnist-cls-4-sym}): class 4 -- MNIST under $\eta = 0.7$ (symmetric) noise}
	
	\label{fig:mnist-cls-thresh-sym}	
\end{figure*}

\begin{figure*}
	\hspace*{0.8cm}\subfloat[]{\label{fig:mnist-cls-0-cc}\includegraphics[scale=0.04]{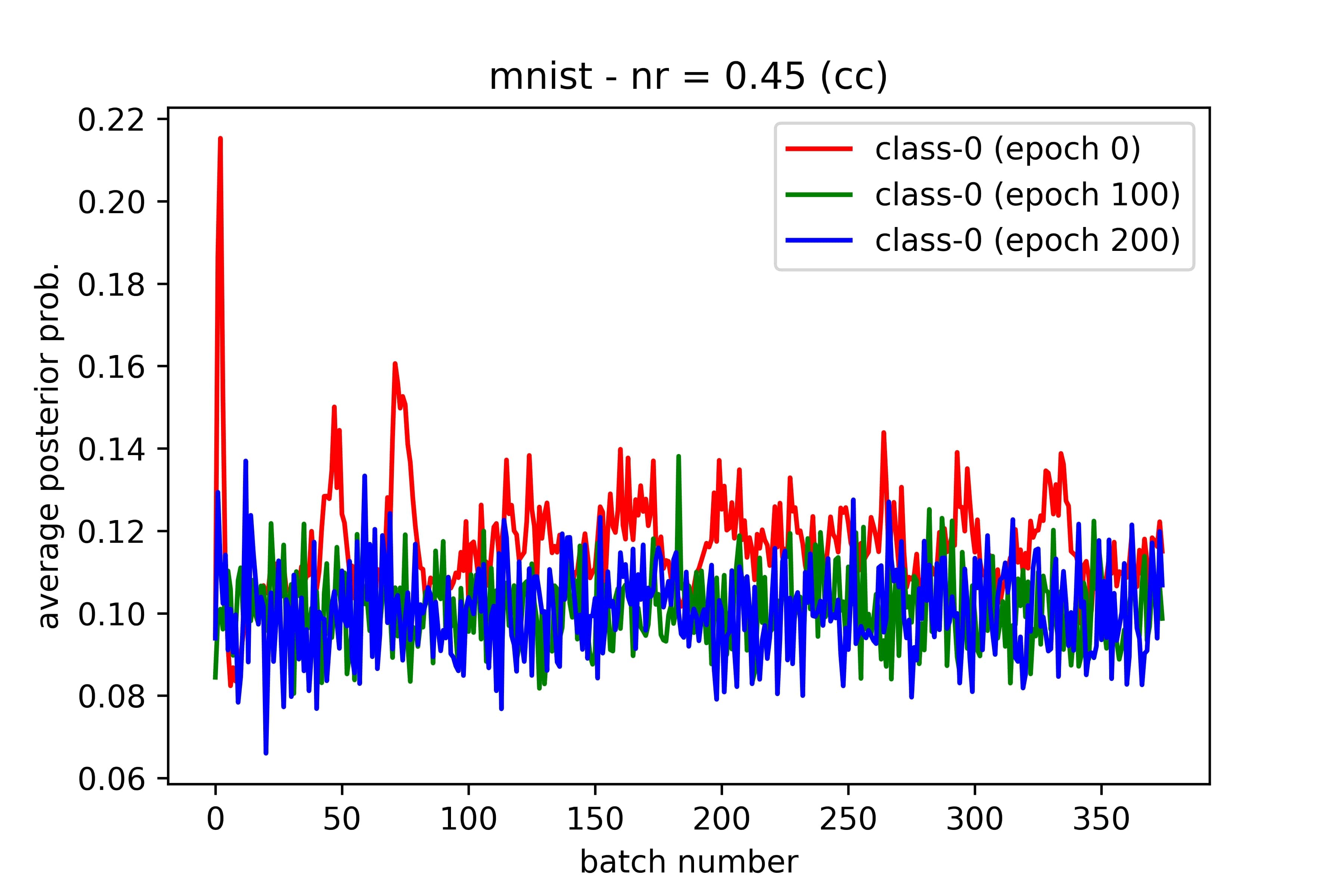}} \subfloat[]{\label{fig:mnist-cls-3-cc}\includegraphics[scale=0.04]{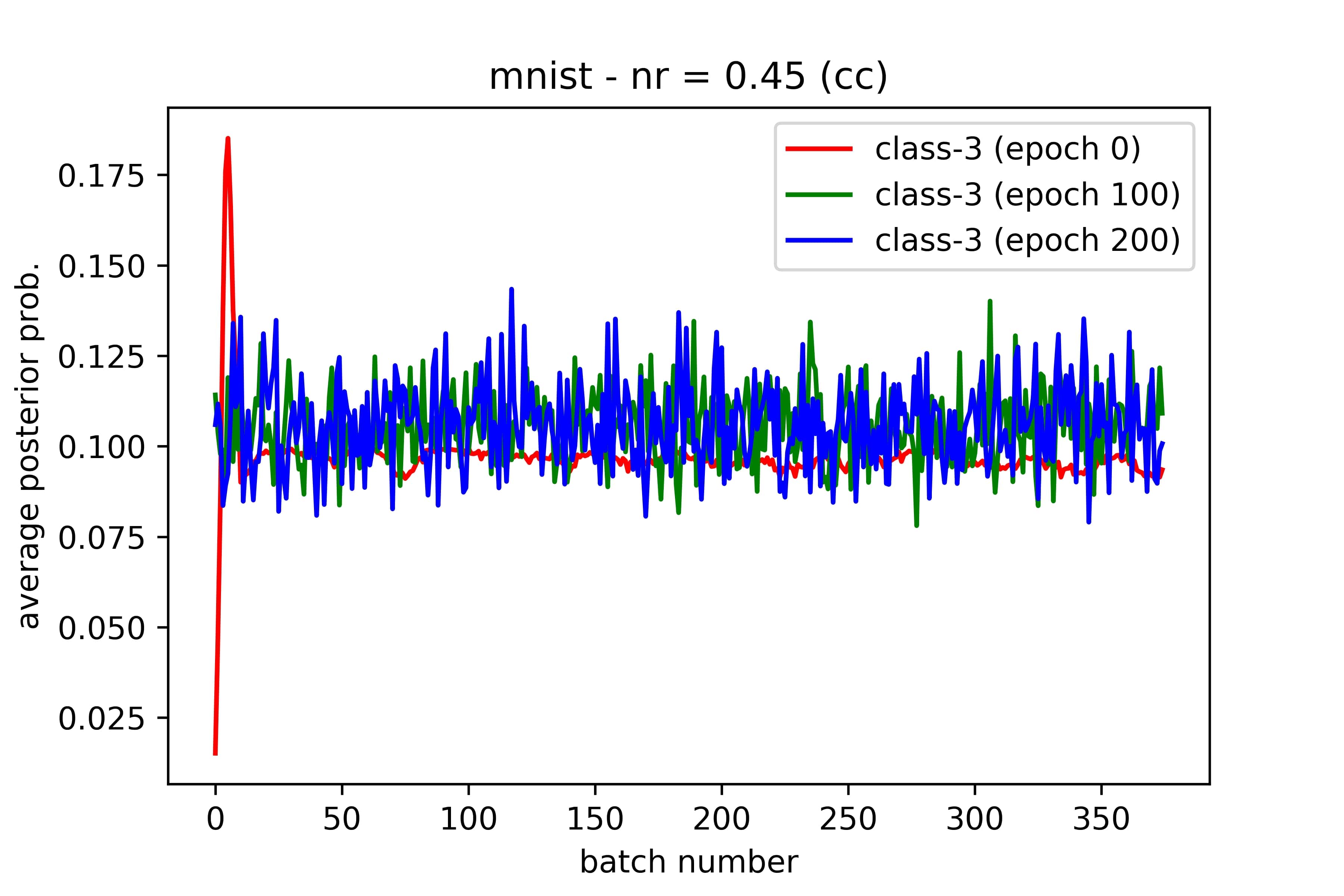}}
	\subfloat[]{\label{fig:mnist-cls-4-cc}\includegraphics[scale=0.04]{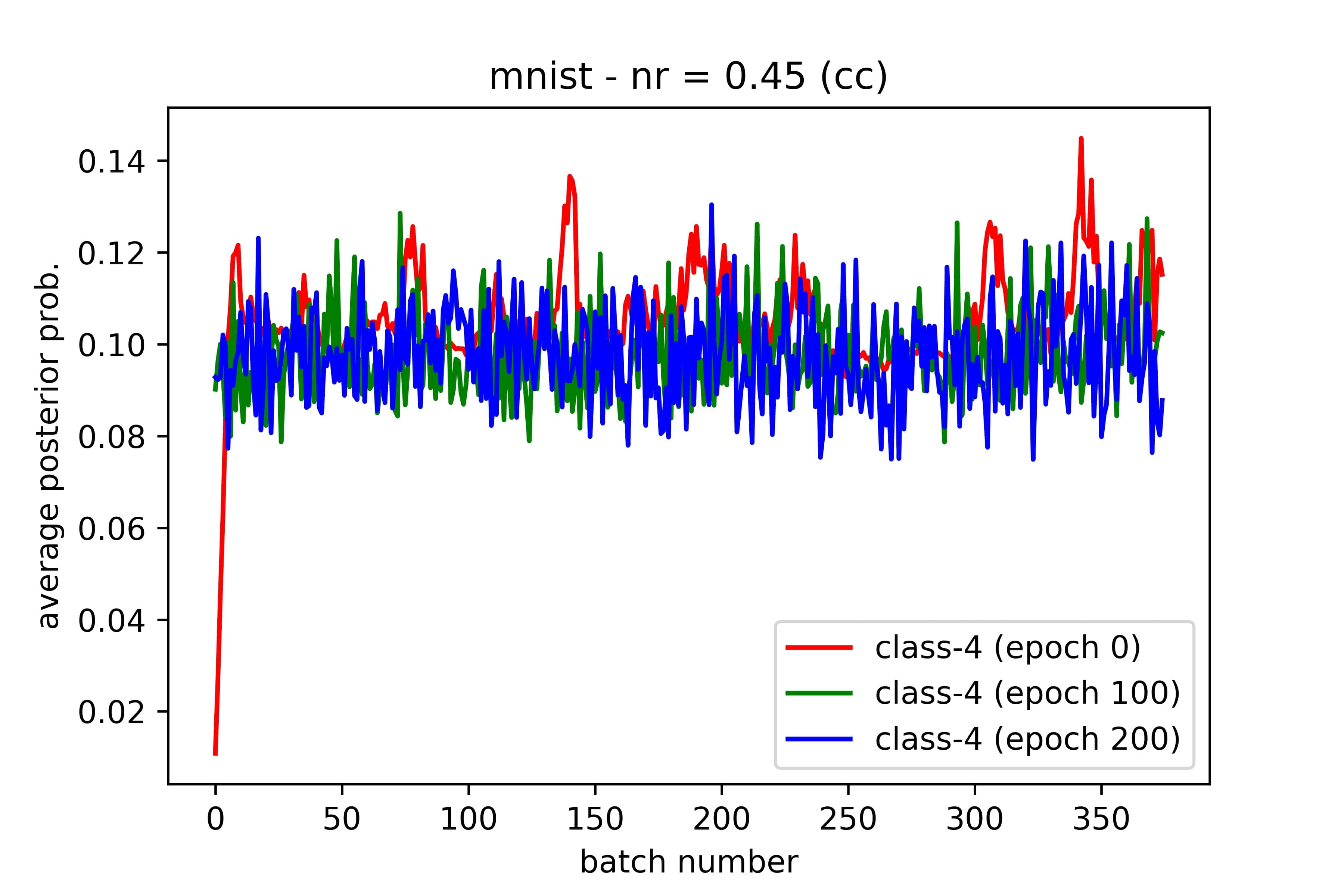}}
	
	\caption{Average class-wise posterior probability across mini-batches; (\subref{fig:mnist-cls-0-cc}): class 0; (\subref{fig:mnist-cls-3-cc}): class 3; (\subref{fig:mnist-cls-4-cc}): class 4 -- MNIST under $\eta = 0.45$ (class-conditional) noise}
	
	\label{fig:mnist-cls-thresh-cc}	
\end{figure*}

\begin{figure*}
	\hspace*{0.8cm}\subfloat[]{\label{fig:cifar-cls-7-sym}\includegraphics[scale=0.04]{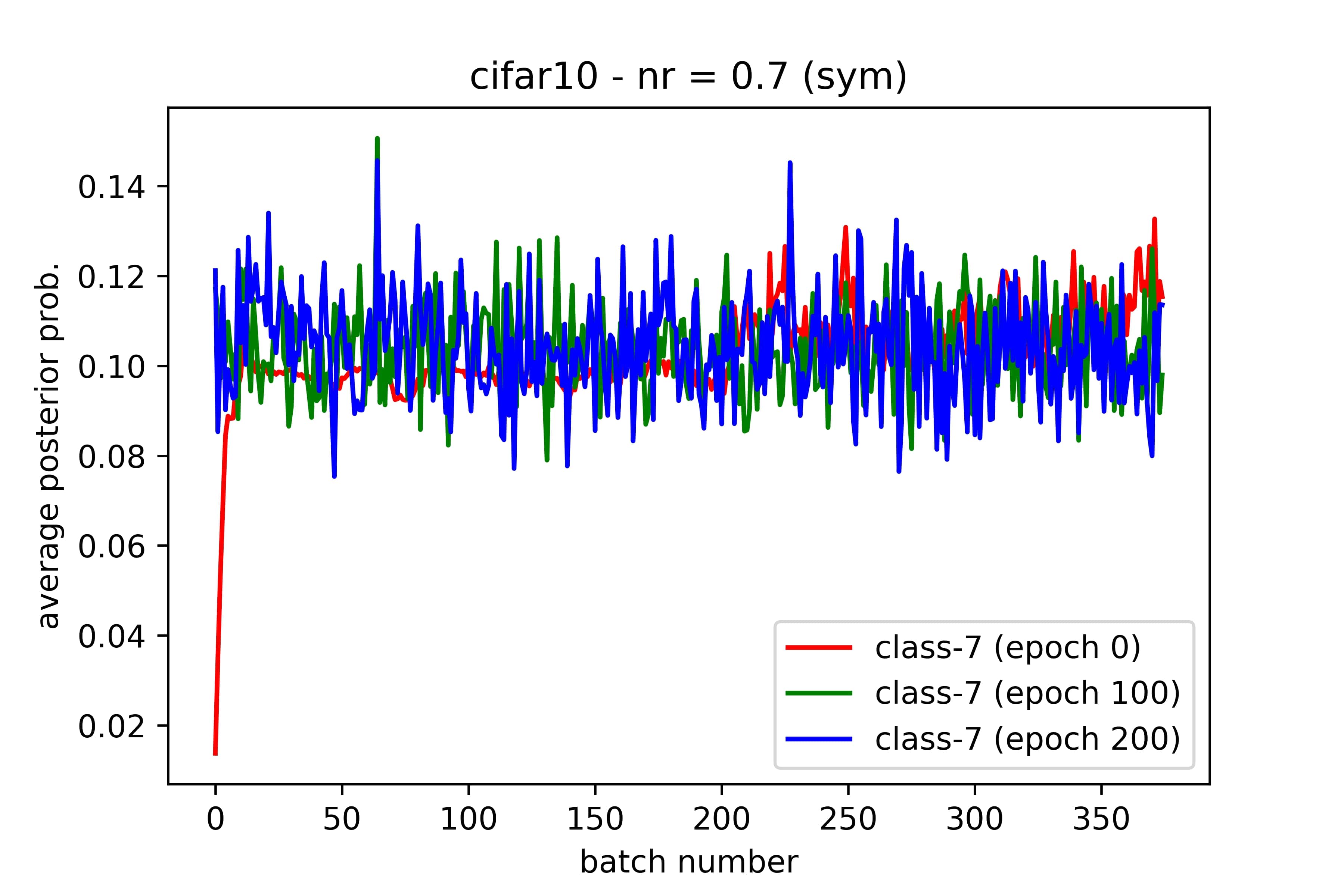}} \subfloat[]{\label{fig:cifar-cls-8-sym}\includegraphics[scale=0.04]{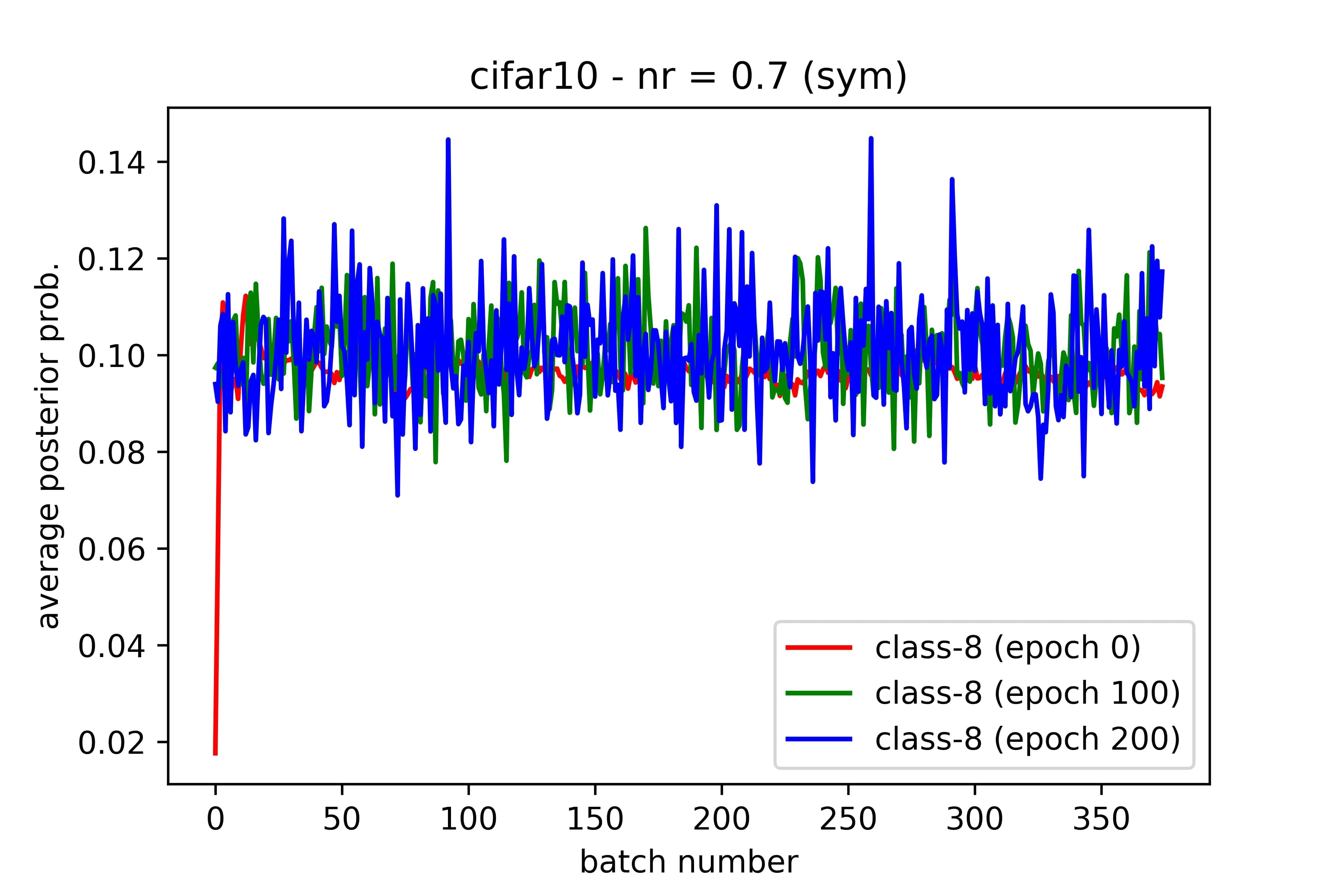}}
	\subfloat[]{\label{fig:cifar-cls-9-sym}\includegraphics[scale=0.04]{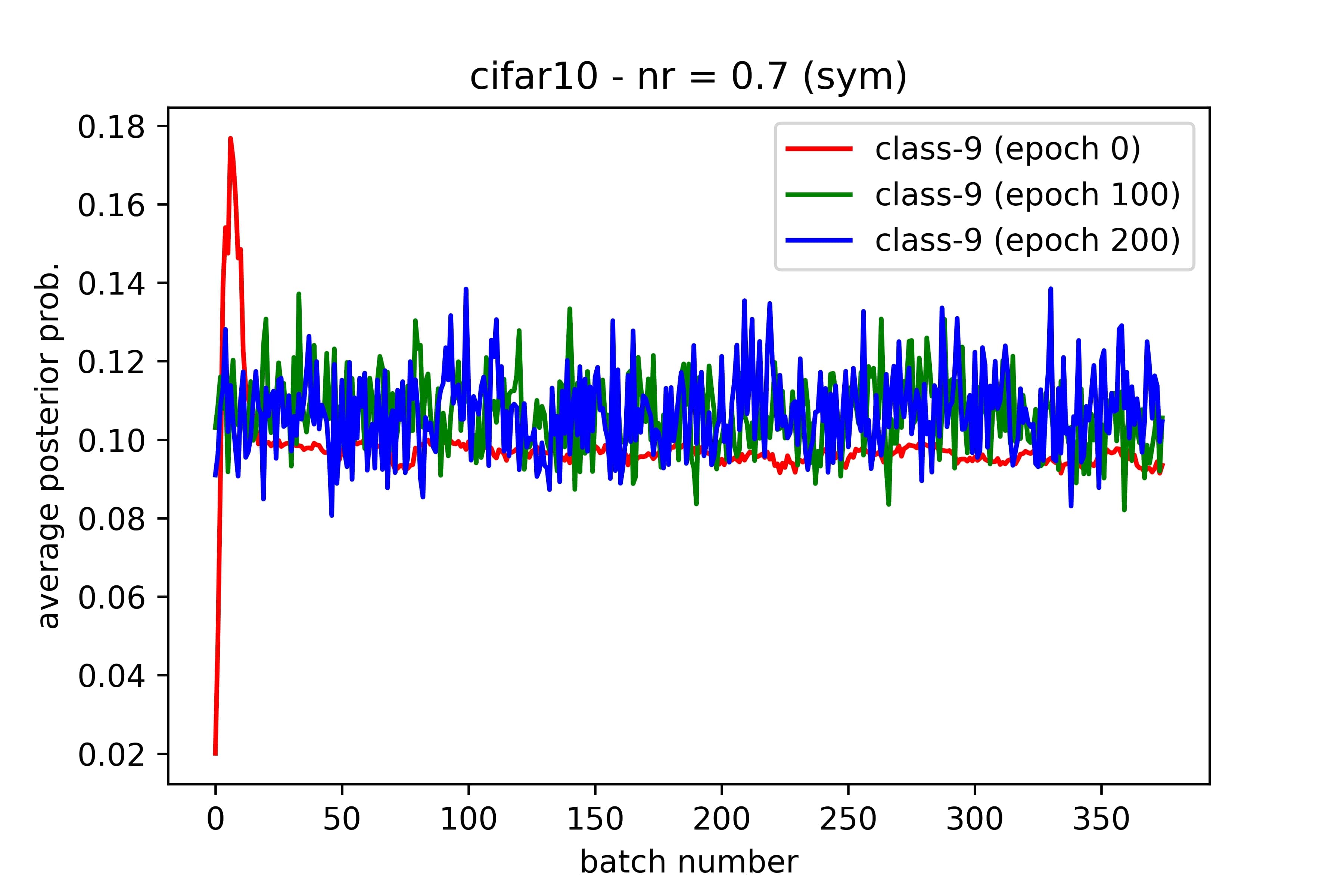}}

	\caption{Average class-wise posterior probability across mini-batches; (\subref{fig:cifar-cls-7-sym}): class 7; (\subref{fig:cifar-cls-8-sym}): class 8; (\subref{fig:cifar-cls-9-sym}): class 9 -- CIFAR-10 under $\eta = 0.7$ (symmetric) noise}
	
	\label{fig:cifar-cls-thresh-sym}	
\end{figure*}

%\vspace*{-20cm}

\begin{figure*}
%	\vspace*{-2cm}
	\hspace*{0.8cm}\subfloat[]{\label{fig:cifar-cls-6-cc}\includegraphics[scale=0.04]{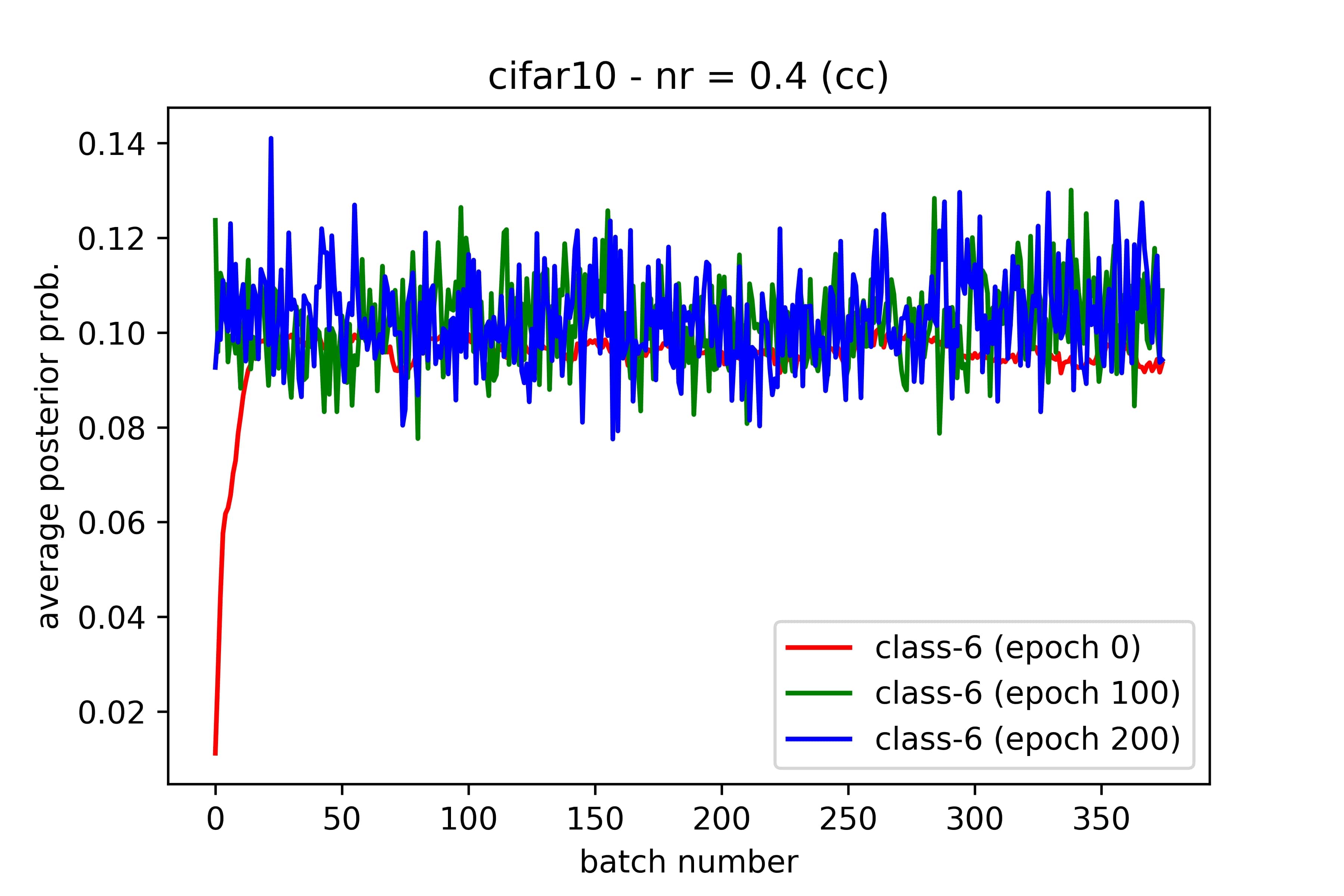}} \subfloat[]{\label{fig:cifar-cls-7-cc}\includegraphics[scale=0.04]{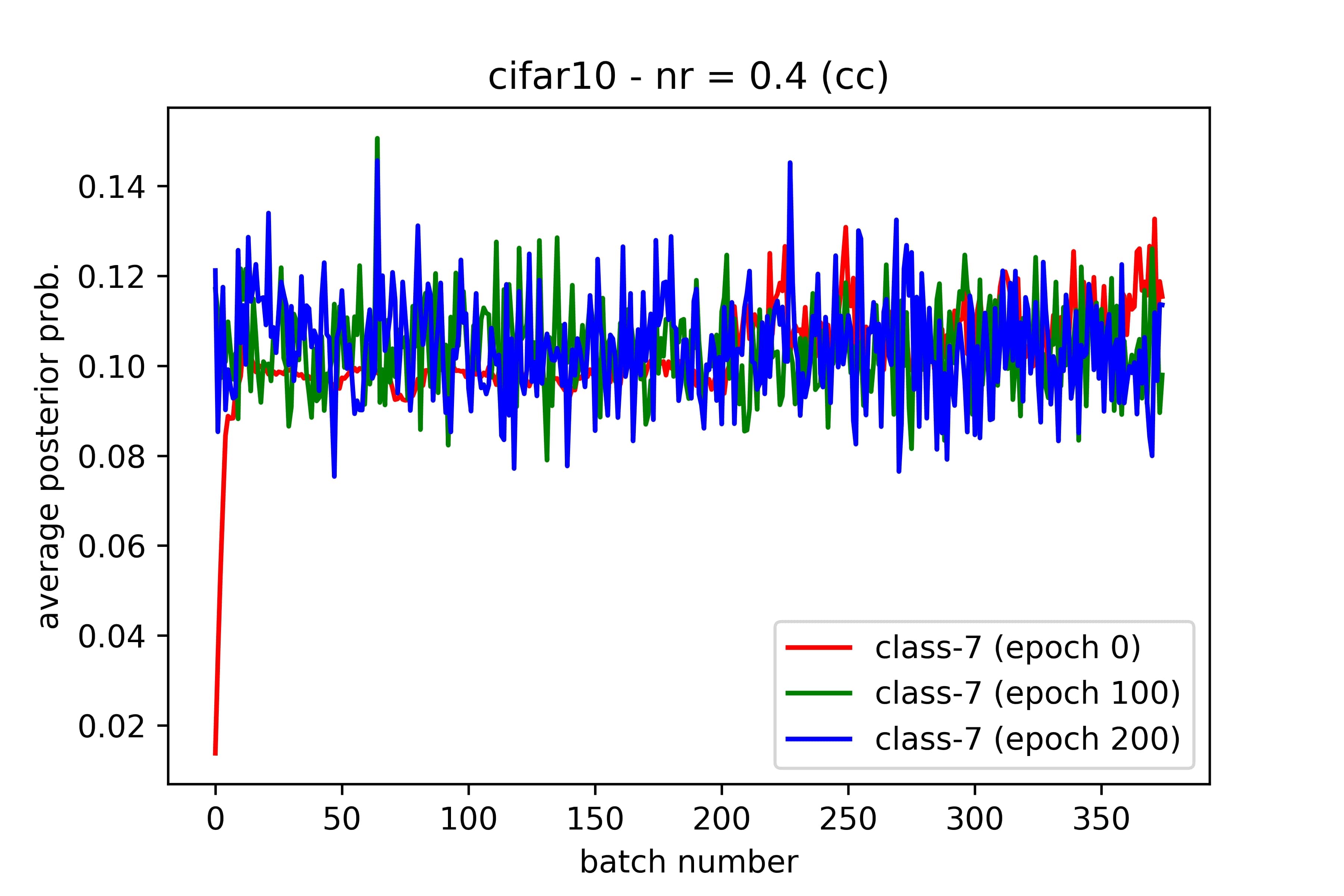}}
	\subfloat[]{\label{fig:cifar-cls-8-cc}\includegraphics[scale=0.04]{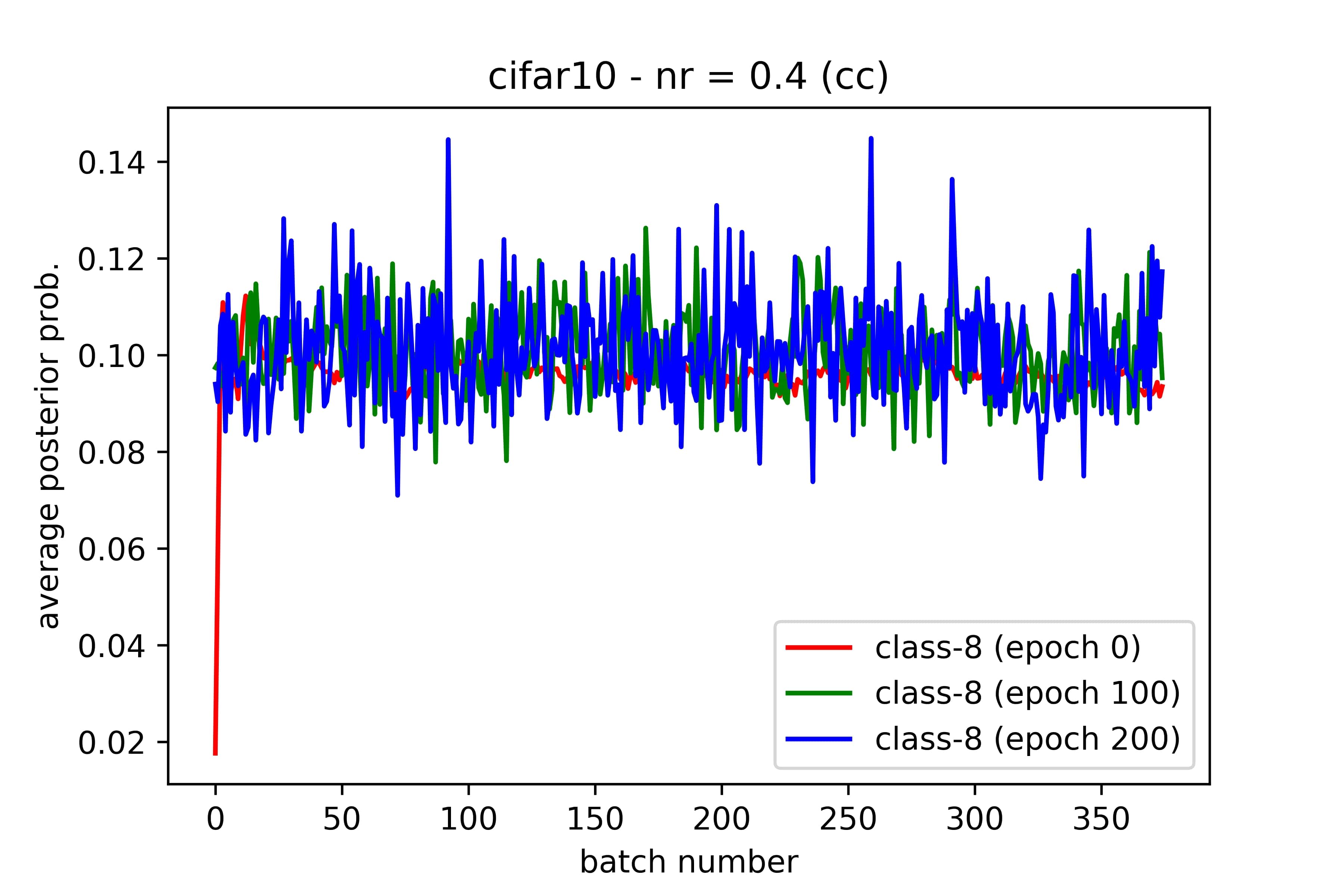}}

	\caption{Average class-wise posterior probability across mini-batches; (\subref{fig:cifar-cls-6-cc}): class 6; (\subref{fig:cifar-cls-7-cc}): class 7; (\subref{fig:cifar-cls-8-cc}): class 8 -- CIFAR-10 under $\eta = 0.4$ (class-conditional) noise}
	
	\label{fig:cifar-cls-thresh-cc}	
\end{figure*}

%\clearpage

\end{document}